\documentclass{article}

\PassOptionsToPackage{numbers}{natbib}
\usepackage[eandd,preprint]{neurips_2026}

\usepackage{amsmath,amsfonts,bm}

\def\eqref#1{equation~\ref{#1}}

\def\1{\bm{1}}

\DeclareMathAlphabet{\mathsfit}{\encodingdefault}{\sfdefault}{m}{sl}
\SetMathAlphabet{\mathsfit}{bold}{\encodingdefault}{\sfdefault}{bx}{n}

\usepackage{amsmath}
\usepackage{amssymb}
\usepackage{amsfonts}
\usepackage{mathtools}
\usepackage{amsthm}
\usepackage[utf8]{inputenc}
\usepackage[T1]{fontenc}
\usepackage{hyperref}
\hypersetup{hidelinks}
\usepackage{url}
\usepackage{microtype}
\usepackage{nicefrac}
\usepackage{graphicx}
\usepackage{subcaption}
\usepackage{booktabs}
\usepackage{makecell}
\usepackage{multirow}
\usepackage{array}
\usepackage{xcolor}
\usepackage{colortbl}
\usepackage{adjustbox}
\usepackage[capitalize,noabbrev]{cleveref}
\definecolor{lightgray}{gray}{0.9}

\usepackage[most]{tcolorbox}
\definecolor{promptAccent}{RGB}{15,56,112}
\definecolor{promptBack}{RGB}{245,248,255}

\newtcolorbox[auto counter]{PromptBox}[2][]{%
  enhanced,
  breakable,
  colback=promptBack,
  colframe=promptAccent,
  coltitle=white,
  colbacktitle=promptAccent,
  fonttitle=\bfseries\small\ttfamily,
  fontupper=\ttfamily\footnotesize,
  title={\ifx&#2&Prompt~\thetcbcounter\else#2\fi},
  boxed title style={
    boxrule=0pt,
    top=2pt,
    bottom=2pt,
    left=6pt,
    right=6pt
  },
  attach boxed title to top left={yshift=-1.5mm, xshift=4mm},
  boxrule=0.75pt,
  arc=2pt,
  boxsep=5pt,
  before skip=6pt,
  after skip=6pt,
  borderline west={2pt}{0pt}{promptAccent},
  drop shadow southeast,
  sharp corners=south,
  before upper={\vspace{2pt}},
  #1
}

\theoremstyle{plain}

\theoremstyle{definition}

\theoremstyle{remark}

\title{SANEval: Open-Vocabulary Compositional Benchmarks with Failure-mode Diagnosis}

\author{%
  Rishav Pramanik\textsuperscript{1}\thanks{These authors contributed equally to the work}~~~~Ian Nielsen~\textsuperscript{2$*$}~~~~Jeff Smith~\textsuperscript{3}~~~~Saurav Pandit~\textsuperscript{4} \\ \textbf{Ravi P. Ramachandran~\textsuperscript{2}~~~~Zhaozheng Yin\textsuperscript{1}}\\
  \textsuperscript{1}Stony Brook University, New York, USA\\
  \textsuperscript{2}Rowan University, New Jersey, USA\\
  \textsuperscript{3}C Proof\\
  \textsuperscript{4}Prelight AI\\
}

\begin{document}

\maketitle

\begin{abstract}
  Text-to-image (T2I) models still struggle with compositional prompts that require multiple objects, correctly bound attributes, accurate counts, or specific spatial relations. Some evaluation benchmarks remain constrained by fixed detector vocabularies and aggregated scores, while human-validation protocols vary. We introduce \textbf{SANEval} (Spatial, Attribute, and Numeracy Evaluation), an open-vocabulary compositional benchmark with two contributions: (i) a public dataset of $\sim$$5{,}250$ prompts paired with images from six state-of-the-art T2I models, partitioned into a Simple split (algorithmically generated, human-validated) and a Hard split (fully human-written), and labeled by compositional task type (spatial, numeracy, color, shape, texture), and (ii) a modular evaluation pipeline that combines an LLM-based prompt parser, an open-vocabulary detector with LLM-driven synonym expansion and mapping, and three scorers for attribute binding, spatial relations, and numeracy. Each scorer emits both a numerical score and structured diagnostic feedback identifying what is missing, extra, or mis-bound. We validate SANEval against human ratings ($n{=}500$, $4$ annotators, $5$ categories). On the shared in-vocabulary sample, SANEval has positive Pearson correlations in every category and an overall correlation of $r{=}0.222$, compared with $-0.016$ for CompBench++ and $-0.030$ for GenEval. Using unrounded human means reduces SANEval's correlation to $0.104$, showing that alignment remains modest and depends on how ratings are aggregated. A manual audit of $24{,}012$ detections across $10{,}040$ image-level samples decomposes the pipeline into a perception stage and an LLM/VLM verification stage and measures returned-box precision at $96.58\%$; verification-only errors contribute $1.05\%$ of all assessed boxes, while removing synonym expansion and mapping reduces scores by $43$--$76\%$ in ablation, demonstrating the scores' sensitivity to this module. The dataset is publicly released at \url{https://huggingface.co/datasets/saneval/saneval-release} and the evaluation pipeline at \url{https://github.com/nielseni6/saneval}.
\end{abstract}

\section{Introduction} \label{S:Introduction}
Text-to-image (T2I) generation has advanced rapidly through model families such as Stable Diffusion~\cite{rombach2022highresolution}, Imagen~\cite{saharia2022photorealistic}, GPT~\cite{openai2024gpt4ocard,pramanik2026unlocking}, and Gemini Image (aka Nano Banana)~\cite{Google_2025}, with broad adoption across industries from marketing to entertainment. Yet these systems still struggle with seemingly simple compositional tasks: attribute binding on multiple objects, capturing spatial relationships, or handling numbers. A primary research bottleneck is the lack of reliable evaluation methods, as existing benchmarks remain limited in scope and fail to capture many of these challenges.

Existing T2I evaluation methods fall into three broad families. Embedding-based scores such as CLIPScore~\cite{hessel2021clipscore}, and representations from CLIP~\cite{radford2021learning}, DINO~\cite{caron2021emerging}, and BERT~\cite{devlin2019bert} were not designed for fine-grained composition and can miss object-level errors~\cite{dinh2022tise,ross2024what}. VLM-as-judge approaches such as Davidsonian Scene Graphs~\cite{cho2024davidsonian} and ConceptMix~\cite{wu2024conceptmix} decompose composition into questions, providing evidence that can distinguish missing objects from incorrect attributes. Detector-based benchmarks~\cite{ghosh2023geneval,huang2023compbench,huang2025compbench++} provide structured outputs, but their coverage is constrained by the configured vocabularies of their underlying detectors. Human validation, including TIFA~\cite{hu2023tifa} and GenAI-Bench~\cite{li2024genaibench}, provides an external check on whether automated scores track human perception of compositional fidelity.

Two recent developments make a complementary design possible: open-vocabulary detectors~\cite{wang2025yoloerealtimeseeing} that accept arbitrary class queries, and instruction-following LLMs that parse prompts into structured representations and reason over detector outputs. We combine these into \textbf{SANEval} (Spatial, Attribute, and Numeracy Evaluation), an open-vocabulary compositional benchmark with two parts: an open-source dataset and a modular evaluation pipeline. The dataset comprises SANEval-Simple ($\sim$$5{,}000$ algorithmically generated, human-validated prompts) and SANEval-Hard ($250$ fully human-written prompts), paired with images from six state-of-the-art T2I models and metadata identifying the model, split, and task. The pipeline has a \textit{Prompt Understanding Module} that extracts objects, attributes, relations, and counts; an \textit{Enhanced Object Detection Module} that pairs YOLO-E~\cite{wang2025yoloerealtimeseeing} with LLM-driven synonym expansion and mapping; and three scorers for attribute binding, spatial relations, and numeracy that each emit a score and structured diagnostic feedback identifying missing objects, extra objects, and incorrect bindings.

We validate the benchmark against human ratings ($500$ prompt--image pairs, $4$ annotators, $5$ categories). On the shared in-vocabulary sample, SANEval achieves a positive Pearson correlation in every category and an overall correlation of $r{=}0.222$, compared with $r{=}-0.016$ for CompBench++~\cite{huang2025compbench++} and $r{=}-0.030$ for GenEval~\cite{ghosh2023geneval}. These results support more consistent directional alignment in this sample, while the absolute correlations remain modest. A two-stage manual audit of $24{,}012$ detections across $10{,}040$ image-level samples measures returned-detection \emph{precision} at $96.58\%$. Verification-only errors (synonym mapping and VLM-as-judge attribute reading) account for $1.05\%$ of all assessed boxes, while perception-only errors account for $2.09\%$. Backbone-swap experiments (Gemini Flash 2.5 vs.\ Llama-4) yield an average score deviation of $\sim$$0.04$ across the two evaluated backbones. The reported threshold-sensitivity sweep shows correlation decreasing smoothly in both directions away from the a priori-chosen attribute-binding threshold.

Our contributions are as follows:
\begin{itemize}
    \item \textbf{An open-vocabulary compositional benchmark with diagnostic feedback.} SANEval evaluates T2I models on attribute binding (color, shape, texture), spatial relations, and numeracy without any fixed object vocabulary. Each of three scorers emits both a numerical score and structured per-error annotations identifying what is missing, what is extra, and which attributes fail to match the prompt.
    \item \textbf{A public dataset and reproducible pipeline.} We release $\sim$$5{,}250$ prompts (SANEval-Simple~$+$~SANEval-Hard) paired with images from six state-of-the-art T2I models, labeled by originating model, split, and compositional task type, at \url{https://huggingface.co/datasets/saneval/saneval-release}; the evaluation pipeline source code is at \url{https://github.com/nielseni6/saneval}.
    \item \textbf{Validated against human judgment.} We compare SANEval, CompBench++, and GenEval against human ratings on the same in-vocabulary prompt--image pairs. SANEval has positive Pearson correlations in every category and an overall correlation of $r{=}0.222$, compared with $-0.016$ and $-0.030$ for the two baselines.
    \item \textbf{End-to-end pipeline precision, quantified.} An audit of $24{,}012$ detections across $10{,}040$ image-level samples measures \emph{precision} at $96.58\%$, with verification-only errors accounting for $1.05\%$ of all assessed boxes. Synonym-module ablations reduce scores by $43$--$76\%$; the two-backbone comparison and reported threshold-sensitivity sweep further characterize the pipeline's dependence on its components and settings.
\end{itemize}

\section{Background and Related Work} \label{S:Background}

\paragraph{Compositional T2I evaluation.}
Early T2I evaluation relied on object-centric datasets (CUB Birds~\cite{wah2011birds}, Oxford Flowers~\cite{xia2017inception}, COCO-Captions~\cite{chen2015microsoft}) and image-quality metrics such as FID~\cite{pramanik2025distilling} and CLIPScore~\cite{hessel2021clipscore}, which can miss compositional errors such as incorrect counts, swapped attributes, and misplaced objects. Targeted compositional benchmarks~\cite{cho2023dall,petsiuk2022human,bakr2023hrsbench,huang2023compbench,huang2025compbench++,ghosh2023geneval,xiwei2024ella,li2025easier} address this by decomposing prompt adherence along axes such as spatial relations, numeracy, and attribute binding. However, rule-based matching can be brittle: spatial relations may be extracted via named entity recognition over fixed lexical cues~\cite{srinivasa2018natural}, counting can fail on rare objects (e.g., \textit{capybara}, \textit{albatross}) outside the detector's vocabulary, and attribute binding can conflate attribute--object associations in multi-object prompts.

\paragraph{VQA vs.\ OD verification, and the vocabulary ceiling.}
Existing benchmarks rely on one of two paradigms. VQA-based approaches~\cite{xiwei2024ella,li2025easier,hu2023tifa}, often built on BLIP-VQA~\cite{li2022blip}, query an image and read off yes/no judgments; they handle natural language flexibly and can provide question-level evidence, but their outputs are sensitive to phrasing~\cite{kim2025visual}. OD-based methods~\cite{ghosh2023geneval,huang2025compbench++,zang2025contextual} produce bounding boxes and labels for direct geometric and counting reasoning, but inherit their detectors' closed vocabularies. These include MS-COCO classes for the configured Mask2Former~\cite{cheng2022mask2former}, multiple dataset taxonomies for UniDet~\cite{zhou2022simple}, and $20$ classes for PASCAL-VOC-trained models~\cite{pramanik2024masked}. Objects outside the configured taxonomy cannot be evaluated directly. Recent open-vocabulary detectors such as YOLO-E~\cite{wang2025yoloerealtimeseeing} accept arbitrary class queries through a decoupled prompt encoder, removing the vocabulary ceiling while retaining structured outputs.

\paragraph{Where SANEval fits.}
SANEval takes the hybrid path: it pairs an open-vocabulary detector with LLM-driven prompt parsing, synonym expansion, and synonym mapping, retaining OD's structured outputs while removing the closed-vocabulary constraint. It also emits structured per-prompt feedback alongside every score and provides a shared-sample human-correlation comparison between three compositional T2I evaluators (Section~\ref{S:Results:Human}).

\section{The SANEval Framework} \label{S:Methodology}
\begin{figure*}[ht!]
\centering
\includegraphics[width=0.85\linewidth,keepaspectratio]{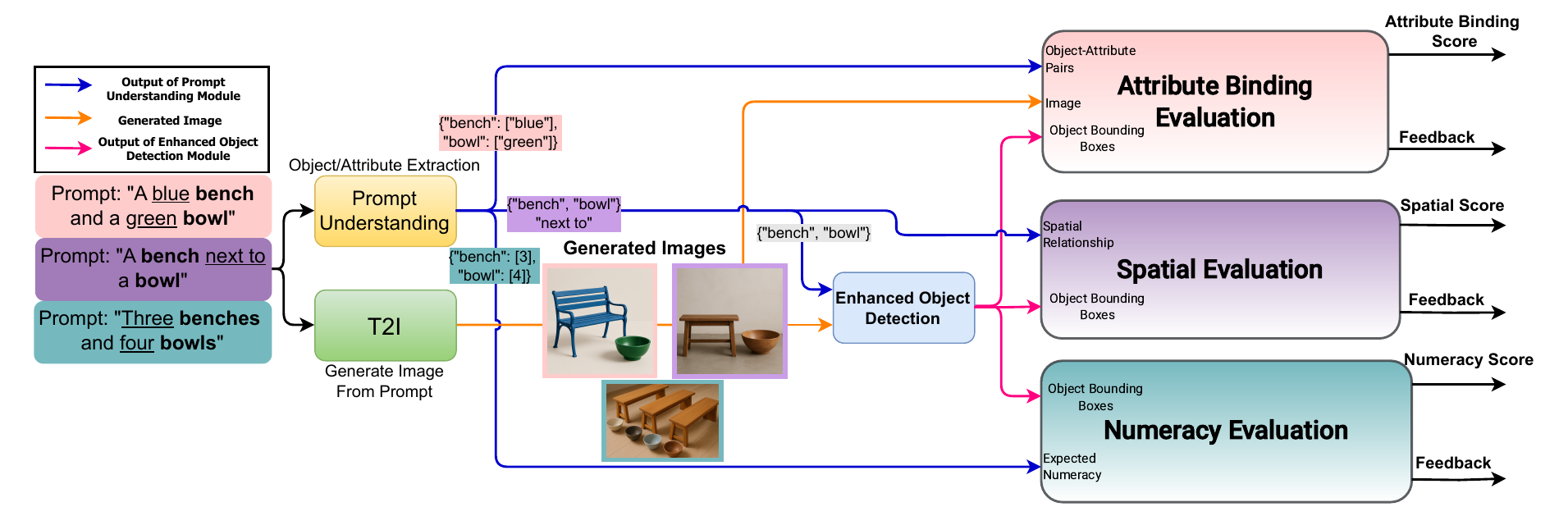}
\caption{Block diagram of the SANEval benchmark.}
\label{fig:saneval_block_diagram}
\end{figure*}

SANEval combines two shared modules, a Prompt Understanding Module (\S\ref{S:Methodology:PromptExtraction}) and an Enhanced Object Detection Module (\S\ref{S:Methodology:ObjectDetection}), with three scorers for attribute binding (\S\ref{S:Methodology:AttributeBinding}), spatial relations (\S\ref{S:Methodology:Spatial}), and numeracy (\S\ref{S:Methodology:Numeracy}). Fig.~\ref{fig:saneval_block_diagram} gives an overview.

\subsection{Dataset Curation} \label{S::Methodology:Dataset}
We curate two prompt sets. \textbf{SANEval-Simple}\footnote{\url{https://huggingface.co/datasets/saneval/saneval-release}} contains $\sim$$5{,}000$ prompts generated deterministically by templating curated lists of objects, attributes, number words, and spatial relations. An author reviewed each prompt alongside its generated image using the human-rating interface (Figures~\ref{fig:human_spatial}--\ref{fig:human_texture}), with essentially no prompts requiring removal or replacement. Prompts are paired with images from six state-of-the-art T2I models. The global prompt counts are $1{,}122$, $2{,}127$, $430$, $394$, and $402$ for one to five objects, $75$ at each count from six to ten, and $30$ at each count from eleven to fifteen. \textbf{SANEval-Hard} adds $250$ fully human-written prompts featuring metaphors, dense scenes, and unusual compositions.

The category breakdown is as follows. \emph{Attribute binding}: $3{,}000$ Simple prompts ($1{,}000$ each for color, shape, texture) plus $150$ Hard ($50$ per type, with every mentioned object carrying an attribute). \emph{Spatial relationships}: $1{,}000$ Simple plus $50$ Hard (two objects with spatial keywords appearing anywhere in the sentence). \emph{Numeracy}: $1{,}000$ Simple plus $50$ Hard (specific cardinality for every mentioned object). Each released record identifies the originating model, split (\texttt{simple}/\texttt{hard}), and compositional task (\texttt{spatial}, \texttt{numeracy}, \texttt{color}, \texttt{shape}, \texttt{texture}); Appendix~\ref{AS.schema} documents the full schema. The public release contains $29{,}998$ image records after failed generations are excluded.

\subsection{Prompt Understanding Module} \label{S:Methodology:PromptExtraction}
We employ an LLM guided by structured system prompts (Appendix~\ref{AS.prompts}) to parse each input into objects, attributes, spatial relations, and counts. The output is a set of key--value pairs (object $\rightarrow$ list of attributes) plus relation triplets \texttt{<obj\_1, relation, obj\_2>} (e.g., \texttt{<dog, left\_of, cat>}) and per-object counts. The parser is decoupled across scorers: for the prompt \emph{``4 green chairs and 2 red chairs''}, the numeracy scorer sees the color-agnostic aggregate \texttt{\{chair: 6\}}, while the attribute-binding scorer uses object-keyed attribute entries so each detected chair can be cropped and queried individually. This decoupling aims to prevent attribute-aware grouping from inflating numeric counts; however, the released object-keyed attribute aggregation may collapse repeated-instance distinctions.

\subsection{Enhanced Object Detection Module} \label{S:Methodology:ObjectDetection}
\begin{figure*}[ht!]
\centering
\includegraphics[width=0.85\textwidth]{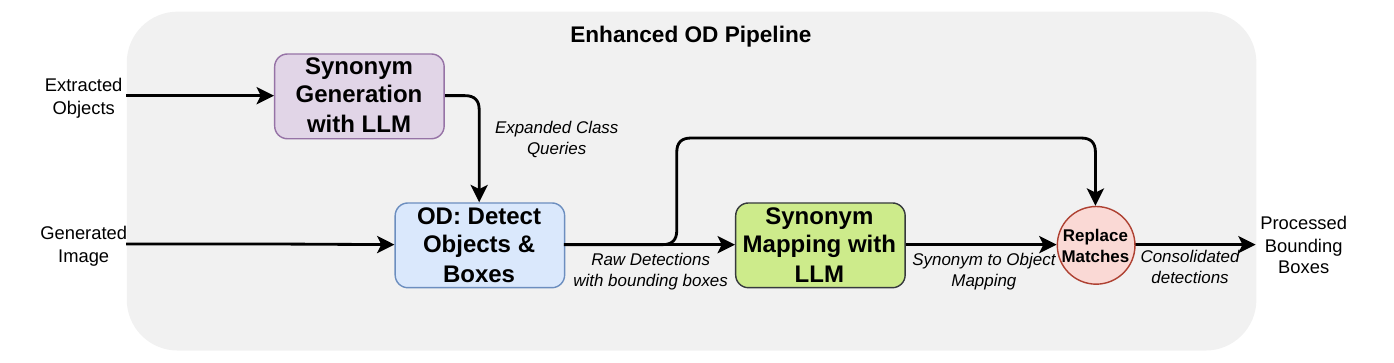}
\caption{Block diagram of the enhanced object detection (OD) pipeline. Given a set of expected objects from the Prompt Understanding Module, the pipeline first generates synonyms and hypernyms via an LLM (e.g., albatross $\rightarrow$ \{albatross, seabird, large bird\}), which are used as expanded class queries for the open-vocabulary detector (YOLO-E). The detector's raw outputs are then passed to an LLM-based synonym mapper, which resolves detected labels back to the original expected objects (e.g., detected ``bird'' $\rightarrow$ expected ``albatross''). This hybrid approach bridges the vocabulary gap between natural language prompts and detector class labels, expanding the detector query vocabulary.}
\label{fig:saneval_od_pipeline}
\end{figure*}

We adopt YOLO-E~\cite{wang2025yoloerealtimeseeing} as the primary detector for its decoupled prompt encoder, which accepts arbitrary class queries; the framework is modular and can be swapped to any open-vocabulary detector. From the expected-object list produced by the Prompt Understanding Module (Section~\ref{S:Methodology:PromptExtraction}), an LLM expands each expected object into a synonym/hypernym set (e.g., \texttt{albatross} $\rightarrow$ \{\texttt{albatross}, \texttt{seabird}, \texttt{large bird}\}); the union of these sets is fed as the detector's class query. YOLO-E's raw detections are then resolved back to the expected objects by an LLM-based synonym mapper that normalizes detected label names (e.g., \texttt{stool} vs.\ \texttt{bench}). This hybrid design (Fig.~\ref{fig:saneval_od_pipeline}) retains OD's structured outputs while expanding lexical coverage. Broad-label mapping alone does not establish fine-grained identity.

\subsection{Attribute Binding Evaluation} \label{S:Methodology:AttributeBinding}
\begin{figure*}[ht!]
    \centering
    \includegraphics[width=0.8\linewidth,keepaspectratio]{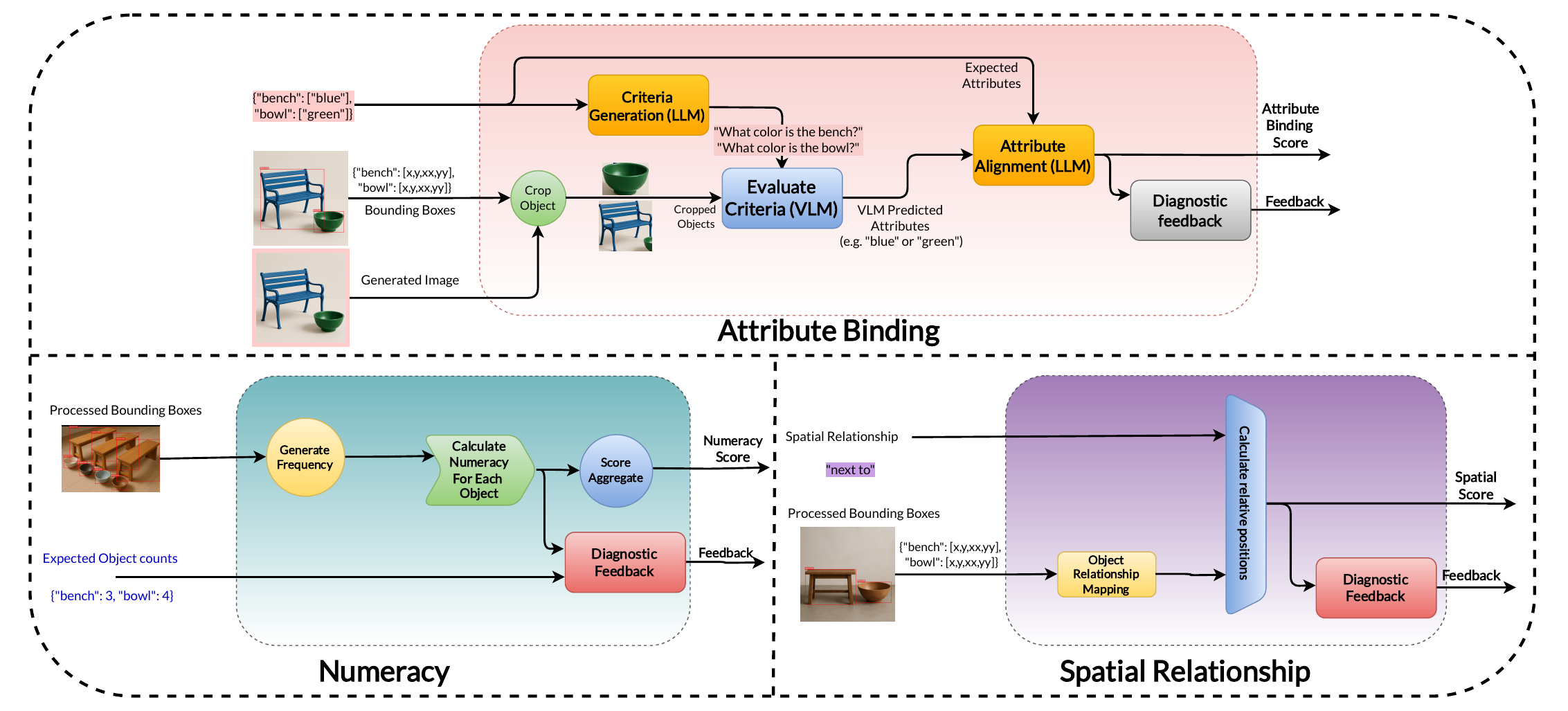}
    \caption{Block diagram descriptions of the three different scorers for the SANEval benchmark.}
    \label{fig:saneval_scorers}
\end{figure*}

The attribute-binding scorer evaluates color, shape, and texture associations (Fig.~\ref{fig:saneval_scorers}). It uses the same Prompt Understanding and Enhanced OD modules as the spatial and numeracy scorers (Sections~\ref{S:Methodology:Spatial}--\ref{S:Methodology:Numeracy}).
\begin{figure*}[ht!]
\centering
\includegraphics[width=0.85\textwidth,keepaspectratio]{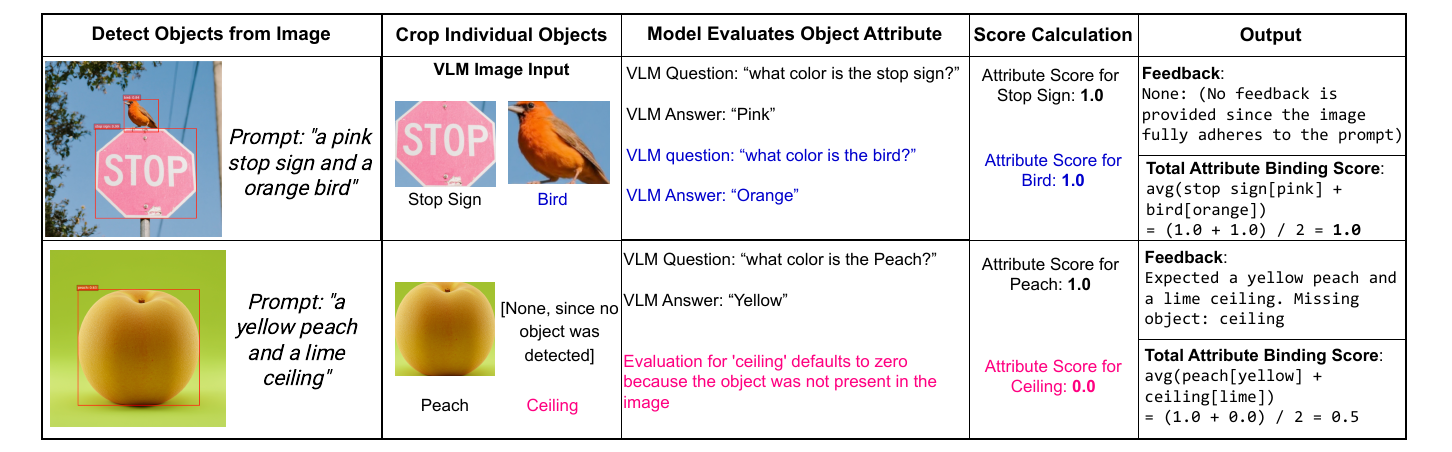}
\caption{Example of SANEval's attribute-binding pipeline. Objects are detected and cropped, then a VLM reads their attributes. The upper example satisfies the prompt (``pink stop sign and orange bird''), yielding a score of 1.0 with no feedback. In the lower example (``yellow peach and a lime ceiling''), the ceiling is missing, yielding a score of 0.5 and corresponding diagnostic feedback.}
\label{fig:attrbind_example}
\end{figure*}
For each detected object, we crop the bounding-box region and pass the crop to a VLM-as-judge (Gemini 2.5 Flash by default), prompted with a targeted, attribute-specific question of the form \texttt{``What \{attribute\_type\} is the \{object\}?''} (e.g., \emph{``What color is the dress?''}). This phrasing avoids leaking the expected attribute into the question while anchoring the query to the object under evaluation. The VLM's free-form answer is then compared against the expected attribute by an LLM-based similarity scorer that returns a continuous score in $[0, 1]$, granting partial credit for near matches (\emph{blue} vs.\ \emph{turquoise} scores between $0.5$ and $1$) and low scores for mismatches (\emph{beige} vs.\ \emph{gray} closer to $0$). The score averages requested attributes per crop, then crops per object-name entry, then expected entries; an undetected entry contributes zero.

Beyond the numerical score, the module emits structured feedback for every flagged binding, e.g.\ \emph{``Poor attribute binding: expected beige dress, but detected gray dress''} or \emph{``Partial binding: expected blue sphere, but detected turquoise sphere.''} If a per-object score falls below the released diagnostic threshold of $0.75$, the binding is flagged as poor alongside the expected value. This two-stage design supplies both a continuous score for benchmarking and structured per-error annotations; the released flag threshold is independent of the scalar score.

\subsection{Spatial Relationship Evaluation} \label{S:Methodology:Spatial}
The spatial scorer evaluates relative positioning using left/right, top/bottom, and proximity cases. We adapt the high-level decomposition of~\cite{huang2025compbench++}, replacing its NER-over-fixed-cues relation extraction with the structured outputs of the Prompt Understanding Module and its closed-vocabulary detection with the Enhanced OD Module. These substitutions accommodate more varied phrasings and object queries. For each prompt-specified relation, we apply a geometric function over the corresponding bounding-box centroids (e.g., does the centroid of object $A$ lie to the left of object $B$). The scorer additionally emits structured feedback such as ``Expected cat to the right of dog: missing object cat.''

Centroid-based 2D evaluation cannot resolve depth-dependent relations (\emph{behind}, \emph{in front of}) from a single image. We leave depth-aware spatial evaluation to future work.

\subsection{Numeracy Evaluation} \label{S:Methodology:Numeracy}
The numeracy scorer compares per-object target counts (from the Prompt Understanding Module) against detected counts (from the Enhanced OD Module): $1.0$ for an exact match, $0.5$ if the object appears with an incorrect count, $0.0$ if it is missing; the prompt-image score is the mean across requested object types. Feedback is structured to report over- and under-generation explicitly (e.g., for the prompt \emph{``three suitcases and two people''} with five suitcases and one person detected: \emph{``two extra suitcases detected; one person missing''}).

\begin{table*}[ht!]
    \centering
    \caption{Combined results for SANEval-Simple benchmarks (Score) and difference with Llama 4 Maverick ($\Delta=s_{\mathrm{Llama}}-s_{\mathrm{Gemini}}$).}
    \label{tab:combinedsaneval_diff}
    \begin{adjustbox}{max width=0.9\linewidth}
        \begin{tabular}{l|cc|cc|cccccc|cc}
            \toprule
            \multirow{3}{*}{Model} & \multicolumn{2}{c|}{\multirow{2}{*}{Spatial}} & \multicolumn{2}{c|}{\multirow{2}{*}{Numeracy}} & \multicolumn{6}{c|}{Attribute-Binding} & \multicolumn{2}{c}{\multirow{2}{*}{Averaged}} \\
            \cmidrule(lr){6-11}
             & & & & & \multicolumn{2}{c}{Color} & \multicolumn{2}{c}{Shape} & \multicolumn{2}{c|}{Texture} & & \\
             & Score & $\Delta$ & Score & $\Delta$ & Score & $\Delta$ & Score & $\Delta$ & Score & $\Delta$ & Score & $\Delta$ \\
            \midrule
            Imagen 3.0       & 0.3524 & -0.050 & 0.5779 & 0.013  & 0.5780 & -0.068 & 0.2499 & -0.040 & 0.4272 & -0.052 & 0.4371 & {-0.039} \\
            Imagen 4.0       & 0.3362 & -0.049 & 0.5192 & 0.005  & 0.5453 & -0.066 & 0.2704 & -0.056 & 0.4339 & -0.052 & 0.4210 & {-0.044} \\
            Imagen 4.0 Ultra & 0.4222 & -0.104 & \textbf{0.6087} & 0.018  & 0.6090 & -0.060 & \textbf{0.3521} & -0.057 & \textbf{0.5041} & -0.057 & \textbf{0.4992} & {-0.052} \\
            Nano Banana      & 0.4076 & -0.094 & \textbf{0.6087} & -0.004 & 0.5927 & -0.068 & 0.2827 & -0.053 & 0.4800 & -0.060 & 0.4743 & {-0.056} \\
            Seedream 3.0     & \textbf{0.4636} & -0.083 & 0.5900 & 0.003  & \textbf{0.6334} & -0.052 & 0.2805 & -0.034 & 0.4969 & -0.049 & 0.4929 & {-0.043} \\
            GPT Image 1      & 0.4372 & -0.105 & 0.5966 & 0.020  & 0.5850 & -0.044 & 0.2972 & -0.028 & 0.4655 & -0.046 & 0.4763 & {-0.041} \\
            \bottomrule
        \end{tabular}
    \end{adjustbox}
\end{table*}
\section{Results and Analysis} \label{S:Results}

\subsection{Experiments on SANEval-Simple}
We evaluate six state-of-the-art T2I generators on SANEval-Simple (Table~\ref{tab:combinedsaneval_diff}). No single model dominates: Imagen 4.0 Ultra leads on the averaged score (driven by numeracy, shape, and texture), Seedream 3.0 follows closely and leads on spatial reasoning and color binding, and Nano Banana matches the top numeracy score. Shape binding is the weakest axis across all models. To assess dependence on the LLM/VLM backbone, we repeated the full evaluation with Llama-4 in place of Gemini Flash 2.5 ($\Delta$ columns). The mean absolute deviation is $0.046$. Llama-4 generally assigns lower scores, although close rankings can change and spatial deviations reach $0.105$. These findings characterize the two evaluated backbones.

\subsection{Experiments on SANEval-Hard}
\begin{table*}[ht!]
    \centering
    \caption{Combined results for SANEval-Hard benchmarks (Score) and difference with Llama 4 Maverick ($\Delta=s_{\mathrm{Llama}}-s_{\mathrm{Gemini}}$).}
    \label{tab:combined_hard_results}
    \begin{adjustbox}{max width=0.9\linewidth}
    \begin{tabular}{l|cc|cc|cccccc|cc}
        \toprule
        \multirow{3}{*}{Model} & \multicolumn{2}{c|}{\multirow{2}{*}{Spatial}} & \multicolumn{2}{c|}{\multirow{2}{*}{Numeracy}} & \multicolumn{6}{c|}{Attribute-Binding} & \multicolumn{2}{c}{\multirow{2}{*}{Averaged}} \\
        \cmidrule(lr){6-11}
         & & & & & \multicolumn{2}{c}{Color} & \multicolumn{2}{c}{Shape} & \multicolumn{2}{c|}{Texture} & & \\
         & Score & $\Delta$ & Score & $\Delta$ & Score & $\Delta$ & Score & $\Delta$ & Score & $\Delta$ & Score & $\Delta$ \\
        \midrule
        Imagen 3.0       & 0.1490 & -0.0502 & 0.3510 & 0.0133  & 0.5030 & -0.068  & 0.2162 & -0.0402 & 0.3841 & -0.0517 & 0.3207 & -0.0394 \\
        Imagen 4.0       & 0.1531 & -0.049  & 0.3294 & 0.0049  & 0.3826 & -0.0655 & 0.2535 & -0.0555 & 0.3430 & -0.0522 & 0.2923 & -0.0435 \\
        Imagen 4.0 Ultra & 0.2462 & -0.104  & 0.3848 & 0.0179  & 0.4944 & -0.0597 & 0.2822 & -0.0574 & 0.4286 & -0.0574 & 0.3672 & -0.0521 \\
        Nano Banana      & 0.2504 & -0.0938 & 0.3660 & -0.0043 & 0.5016 & -0.0683 & \textbf{0.3183} & -0.0527 & \textbf{0.4808} & -0.0599 & 0.3834 & -0.0558 \\
        Seedream 3.0     & 0.2553 & -0.0829 & \textbf{0.4285} & 0.0031  & \textbf{0.5736} & -0.0518 & 0.3031 & -0.0339 & 0.4434 & -0.0485 & \textbf{0.4008} & -0.0428 \\
        GPT Image 1      & \textbf{0.2637} & -0.1051 & 0.3095 & 0.0202  & 0.3097 & -0.0435 & 0.1740 & -0.028  & 0.4125 & -0.0462 & 0.2939 & -0.0405 \\
        \bottomrule
    \end{tabular}
    \end{adjustbox}
\end{table*}
SANEval evaluates the human-written prompts of SANEval-Hard without modification (Table~\ref{tab:combined_hard_results}; visual examples in Fig.~\ref{fig:saneval_examples}). All six models have lower average scores than on SANEval-Simple, and the average leaderboard changes. These differences highlight variation in measured compositional performance across prompt sets and the failure modes that a single-split evaluation can miss. The backbone-swap mean difference is similar on Hard prompts (mean $\Delta = -0.046$).

\subsection{Correlation with Human Judgment} \label{S:Results:Human}
To assess whether SANEval scores reflect human perception of compositional fidelity, four annotators independently rated $500$ prompt--image pairs ($100$ per category: spatial, numeracy, color, shape, texture) on a $1$--$5$ Likert scale. Appendix~\ref{AS.humaneval} describes the protocol, interface, and summary scores (Table~\ref{tab:humaneval} and Figures~\ref{fig:human_spatial}--\ref{fig:human_texture}). We computed per-category Pearson correlations between each evaluator's scores and the rounded mean of the four human ratings. SANEval, CompBench++, and GenEval were evaluated on identical in-vocabulary pairs, excluding unsupported objects.

\begin{table}[ht!]
    \centering
    \caption{Pearson correlation $r$ between automated scores and rounded mean human Likert ratings on the same $500$ in-vocabulary prompt--image pairs ($100$ per category). The final column reports overall correlation rather than the mean of category coefficients. Higher values indicate closer alignment; the best value in each column is bold.}
    \label{tab:human_pearson}
    \begin{adjustbox}{max width=0.95\linewidth}
    \begin{tabular}{lcccccc}
        \toprule
        \textbf{Benchmark} & \textbf{Spatial} & \textbf{Numeracy} & \textbf{Color} & \textbf{Shape} & \textbf{Texture} & \textbf{Overall} \\
        \midrule
        \textbf{SANEval (ours)} & \textbf{0.262} & \textbf{0.241} & \textbf{0.232} & \textbf{0.212} & \textbf{0.125} & \textbf{0.222} \\
        \addlinespace
        CompBench++ & 0.034 & $-0.249$ & 0.013 & 0.006 & 0.113 & $-0.016$ \\
        GenEval & 0.050 & $-0.220$ & 0.066 & 0.001 & 0.080 & $-0.030$ \\
        \bottomrule
    \end{tabular}
    \end{adjustbox}
\end{table}

Table~\ref{tab:human_pearson} shows that SANEval has positive correlations in every category and higher coefficients than both evaluated baselines. Its overall correlation is $0.222$, compared with $-0.016$ for CompBench++ and $-0.030$ for GenEval. Both baselines have negative numeracy correlations and small positive correlations in the remaining categories. Because out-of-vocabulary objects were excluded, vocabulary coverage alone does not explain these differences. The absolute correlations remain modest: the results support more consistent directional alignment on this sample, without establishing strong agreement with human judgment. As a complementary check, prompt-image-level Spearman correlations between SANEval and CompBench++ are uniformly small ($|\rho|\!\le\!0.25$; Appendix~\ref{AS.spearman}), indicating limited agreement between their rankings.

The human-score aggregation also affects the measured alignment. Using unrounded means reduces SANEval's overall Pearson correlation from $0.222$ to $0.104$; CompBench++ and GenEval have $-0.005$ and $-0.027$, respectively. SANEval's Spearman and Kendall correlations with the unrounded means are $\rho=0.075$ ($p=0.091$) and $\tau=0.058$ ($p=0.086$). Overall ordinal Krippendorff's $\alpha=0.848$ indicates high agreement among annotators (Table~\ref{tab:annotator_agreement}), but does not define a Pearson-correlation ceiling.

\subsection{Pipeline Reliability and Error Attribution} \label{S:Results:Reliability}
We quantify the precision of SANEval's returned detections through an audit of two stages: \emph{perception}, which proposes labeled boxes using the open-vocabulary detector, and \emph{LLM/VLM verification}, which includes synonym mapping, VLM-as-judge attribute reading, and LLM similarity scoring. Each incorrect output is attributed to perception, verification, or both.

We manually inspected $24{,}012$ returned detections across $10{,}040$ image-level samples on SANEval-Simple, with $23{,}192$ valid and $820$ incorrect outputs. Four expert annotators (credentials and protocol in Appendix~\ref{AS.audit}) labeled each instance as TP/FP using a custom UI showing the experiment log and bounding-box image side-by-side. Each FP was attributed to perception alone (mislocalized or hallucinated box), LLM/VLM verification alone (correct box but wrong mapper assignment or mis-read attribute), or both.

\begin{table}[ht!]
    \centering
    \caption{Error attribution over $24{,}012$ manually audited detections across $10{,}040$ image-level samples from SANEval-Simple, decomposed by pipeline stage. \emph{Perception} = open-vocabulary detector; \emph{LLM/VLM verification} bundles all downstream LLM- and VLM-based components (synonym mapping, VLM-as-judge attribute reading, and LLM similarity scoring) into a single end-to-end stage. The percentage column reports incorrect returned boxes as a fraction of all $24{,}012$ audited detections (i.e., $100 - 96.58 = 3.42\%$ aggregate). Total row in bold.}
    \label{tab:fpr_attribution}
    \begin{adjustbox}{max width=0.95\linewidth}
    \begin{tabular}{lcc}
        \toprule
        \textbf{Error Source} & \textbf{Count} & \textbf{Incorrect boxes (\%)} \\
        \midrule
        Perception only (open-vocabulary detector)        & 501 & 2.09 \\
        LLM/VLM verification only (synonym mapper or VLM/LLM judge) & 253 & 1.05 \\
        Both perception and verification                  &  66 & 0.27 \\
        \midrule
        \textbf{Total}                                    & \textbf{820} & \textbf{3.42} \\
        \bottomrule
    \end{tabular}
    \end{adjustbox}
\end{table}

The audit measures \emph{precision} of $96.58\%$ on returned detections (Table~\ref{tab:fpr_attribution}). Verification-only and perception-only errors account for $1.05\%$ and $2.09\%$ of all assessed boxes, respectively. Removing synonym expansion and mapping reduces scores by $43$--$76\%$ in the ablation (Table~\ref{tab:syn_ablation}).

Per-object attribute scoring limits the effect of a missed detection: it contributes $0$ for that object while retaining credit for correctly detected objects. Missing $1$ of $3$ distinct expected object entries therefore caps the prompt-level score at $0.67$. Diagnostic feedback (e.g., \emph{``Missing object: ceiling''}) is generated from templates using the same component outputs, without an additional generative step. The audit does not measure missed objects, so returned-detection precision cannot establish recall or missing-object diagnostic accuracy. These require a separate prompt-level audit, which we leave to future work.

\subsection{Sensitivity to Heuristic Choices} \label{S:Results:Sensitivity}
The attribute-binding scorer uses a similarity threshold to flag poor bindings. We chose $0.5$ as the default \emph{before} running the human-correlation study, on the natural a priori grounds that the LLM similarity scorer returns a continuous value in $[0,1]$, so $0.5$ is the midpoint dividing ``more similar than dissimilar'' from ``more dissimilar than similar.'' We then sweep the threshold from $0.3$ to $0.7$ in $0.1$ steps and recompute Pearson correlation with rounded mean human Likert ratings (\S\ref{S:Results:Human}) for each attribute type.

\begin{table}[ht!]
    \centering
    \caption{Sensitivity of SANEval's correlation with human ratings to the attribute-binding similarity threshold. Pearson $r$ against rounded mean human Likert scores; the default ($0.5$) is bolded.}
    \label{tab:threshold_ablation}
    \begin{adjustbox}{max width=0.85\linewidth}
    \begin{tabular}{lcccc}
        \toprule
        \textbf{Threshold} & \textbf{Color (r)} & \textbf{Shape (r)} & \textbf{Texture (r)} & \textbf{Overall (r)} \\
        \midrule
        0.3                       & 0.176 & 0.147 & 0.086 & 0.135 \\
        0.4                       & 0.211 & 0.198 & 0.104 & 0.189 \\
        \textbf{0.5 (ours)}       & \textbf{0.232} & \textbf{0.212} & \textbf{0.125} & \textbf{0.204} \\
        0.6                       & 0.215 & 0.210 & 0.095 & 0.175 \\
        0.7                       & 0.185 & 0.199 & 0.082 & 0.142 \\
        \bottomrule
    \end{tabular}
    \end{adjustbox}
\end{table}

In the reported sweep, the a priori choice of $0.5$ yields the highest correlation for all three attribute types. Correlation decreases smoothly in both directions away from this peak.

The released implementation instead defaults to a $0.75$ diagnostic flag threshold independent of its continuous score; changing that flag alone does not reproduce this reported sweep.

\subsection{Probing the Compositional Limits of Image Generators}
We analyze attribute binding as a function of object cardinality (1, 2, 3--5, 6--10, $>$10 objects per prompt) using SANEval; the full per-model breakdown is reported in Appendix~\ref{AS.cardinality} (Table~\ref{tab:attrbind}). Three patterns are robust across all six models. (i) \emph{Shape binding is the most brittle, color the most robust, texture intermediate}: averaged across models, scores collapse from $0.37$ (1 obj) to $0.13$ ($>$10 obj) for shape, but only from $0.62$ to $0.41$ for color. (ii) \emph{Scores generally decline, with intermediate-bin increases}: Seedream 3.0 and Imagen 4.0 Ultra degrade gracefully (Imagen 4.0 Ultra retains $0.22$ shape accuracy at $>$10 objects, the highest of any model), while Imagen 3.0's shape accuracy falls from $0.32$ to $0.06$. (iii) \emph{Robustness to complexity, not peak performance, is the differentiator}: models cluster tightly on single-object prompts but fan out by 3$\times$ at the highest cardinality.

\subsection{Impact of Enhanced Object Detection Module}
We ablate the LLM-driven synonym expansion and mapping against a surface-form-query baseline that uses the same YOLO-E detector directly with the prompt's surface-form object names (Table~\ref{tab:syn_ablation}). Removing the synonym module collapses scores by $43.23\%$ on numeracy ($0.5835 \rightarrow 0.3312$), $75.69\%$ on spatial ($0.4032 \rightarrow 0.0980$), and $56.94\%$ on attribute binding ($0.4491 \rightarrow 0.1934$); the number of unique categories detected drops in lockstep ($118 \rightarrow 77$, $49 \rightarrow 40$, $136 \rightarrow 128$).
\begin{table}[ht!]
    \centering
    \caption{Ablation demonstrating the importance of LLM-driven synonym expansion (\texttt{syn}) for open-vocabulary reasoning in SANEval. Removing \texttt{syn} leads to substantial performance degradation across all metrics.}
    \label{tab:syn_ablation}
    \begin{adjustbox}{max width=0.9\linewidth}
    \begin{tabular}{lcccccc}
        \toprule
        \multirow{2}{*}{\textbf{Metric}}
        & \multicolumn{2}{c}{\textbf{Score}}
        & \multirow{2}{*}{\textbf{Rel. Drop} $\downarrow$}
        & \multicolumn{2}{c}{\textbf{Unique Objects}} \\
        \cmidrule(lr){2-3} \cmidrule(lr){5-6}
        & \textbf{w/ Syn} & \textbf{w/o Syn} &  & \textbf{w/ Syn} & \textbf{w/o Syn} \\
        \midrule
        Numeracy
        & \textbf{0.5835} & 0.3312
        & \textbf{43.23\%}
        & 118 & 77 \\

        Spatial
        & \textbf{0.4032} & 0.0980
        & \textbf{75.69\%}
        & 49 & 40 \\

        Attribute Binding
        & \textbf{0.4491} & 0.1934
        & \textbf{56.94\%}
        & 136 & 128 \\
        \bottomrule
    \end{tabular}
    \end{adjustbox}
\end{table}

\section{Conclusions} \label{S:Conclusions}
We introduced SANEval, an open-vocabulary compositional benchmark for T2I models that pairs scalar scores with structured per-error feedback. Across six state-of-the-art generators, SANEval measures attribute binding, spatial relations, and numeracy without a fixed object vocabulary, and exposes how each model degrades as compositional complexity grows: shape binding collapses fastest, color is the most robust, and no single model leads across all axes. In our shared in-vocabulary human comparison, SANEval has positive correlations across all five categories and an overall Pearson correlation of $r{=}0.222$, compared with $-0.016$ for CompBench++ and $-0.030$ for GenEval. The alignment remains modest and decreases when unrounded human means are used. An audit of $24{,}012$ detections across $10{,}040$ image-level samples measures returned-box precision at $96.58\%$. Verification-only and perception-only errors account for $1.05\%$ and $2.09\%$ of all assessed boxes, respectively. The backbone comparison and reported threshold sweep characterize sensitivity to the pipeline's settings. Together, the public dataset and modular pipeline support automated scoring and analysis of compositional failures.

\paragraph{Reproducibility and Release.} The SANEval-Simple and SANEval-Hard prompt sets, together with images from all six T2I models evaluated here, are released at \url{https://huggingface.co/datasets/saneval/saneval-release}. Every record carries the originating model, split (\texttt{simple}/\texttt{hard}), and task type (\texttt{spatial}, \texttt{numeracy}, \texttt{color}, \texttt{shape}, \texttt{texture}); see Appendix~\ref{AS.schema} for the full schema. The evaluation pipeline, comprising prompt understanding, enhanced object detection, and the three scorers, is released at \url{https://github.com/nielseni6/saneval}, together with the prompt templates of Appendix~\ref{AS.prompts} and the human-annotation materials of Appendix~\ref{AS.humaneval}. These resources support automated rescoring; original human ratings and audit labels are not included in the image release.

\begin{samepage}
\paragraph{Limitations.} Several limitations remain. First, our spatial evaluation uses 2D bounding-box centroids and cannot resolve depth-dependent relations (\emph{behind}, \emph{in front of}) from a single image. Second, the audit measures returned-box precision, not recall or missing-object diagnostic accuracy (\S\ref{S:Results:Reliability}). Third, SANEval-Hard contains $250$ human-written prompts as a within-task stress test rather than exhaustive coverage; expanding it with diverse compositional patterns is an explicit goal of our community release. Human alignment remains modest and depends on how ratings are aggregated. Both human studies share four experts. Partial-credit similarity lacks independent validation, and only two judge backbones are tested. All six generators are proprietary, including four Google models.
\par
\end{samepage}

\begin{ack}
Rishav Pramanik and Zhaozheng Yin have been supported by NSF grants CMMI-2246673 and CAIF-2331769, and SUNY AI Platform Award.
\end{ack}

\newpage
\appendix
\section*{Appendix}
\section{Use of LLMs}
Large Language Models (LLMs) were employed in a supporting role across different stages of this work. During dataset construction and experimentation, LLMs assisted with coding tasks such as prompt generation, data formatting, and automation scripts, which accelerated the development pipeline but did not replace human design or verification. In addition, LLMs were occasionally used to draft alternative phrasings or refine sections of the manuscript, with the goal of improving readability, flow, and consistency of style.

The core methodological roles of LLMs and VLMs are described in Section~\ref{S:Methodology}. All scientific ideas, experimental design, methodological innovations, and analyses were conceived, implemented, and validated by the authors. Any LLM-generated content, whether in code or text, was subject to careful review, editing, and verification to preserve the accuracy, originality, and integrity of our contributions. Thus, while LLMs served as helpful tools for productivity and clarity, the intellectual and technical content of this paper remains entirely the work of the authors.

\section{Broader Impacts} \label{AS.broader-impacts}
This work introduces an evaluation framework for studying compositional and open-vocabulary behavior in text-to-image generation models. By exposing failure cases that are not covered by some existing benchmarks, it helps clarify where current models succeed and where they fall short, which we view as a positive contribution to transparent and rigorous evaluation of generative systems. The framework is designed for evaluation and analysis only and does not add new generative capabilities, so the most direct potential negative impact is interpretive: SANEval relies on language-based priors, and its outputs may reflect biases present in those models and should be interpreted with care. Overall, the work aims to encourage more careful and transparent evaluation of generative systems rather than to advance generation itself.

\section{Infrastructure and Support}
Unless otherwise noted, both the Vision-Language Model (VLM) and Large Language Model (LLM) components in our framework default to Gemini-2.5-Flash. All experiments were conducted on cloud infrastructure provisioned through Amazon Web Services (AWS). Specifically, we use the Elastic Container Service (ECS) with the Fargate launch type, which allows jobs to be deployed in a serverless fashion on top of pre-configured Docker images. These images were custom-built and reused across multiple runs to ensure consistency, reproducibility, and efficient resource management.

Our compute configuration typically consists of Linux-based operating systems running on the ARM64 architecture, with containers provisioned with 8 virtual CPUs, 12 GB of RAM, and up to 48 GB of disk storage. This setup provides a balanced environment suitable for large-scale evaluation while remaining cost-effective.

For executing inference with VLMs, LLMs, and text-to-image generation models, we rely on API endpoints provided by their respective service providers. In particular, we use endpoints from OpenAI, Google, and ByteDance (via the Fal-AI platform). This API-based design allows us to access state-of-the-art models without the need for extensive local deployment and ensures that our benchmarking pipeline remains modular and easily extendable as newer models become available.

\section{Prompts}
\label{AS.prompts}
This section lists the reported prompt templates used for LLMs and VLMs in our framework. We report them verbatim to ensure clarity and reproducibility. Note that additional post-processing steps such as parsing, schema enforcement, and consistency checks were applied during evaluation, but those details are beyond the scope of this section.

\begin{PromptBox}[]{Prompt 1: Used for generating objects from a prompt}
\label{prompt:obj-prompt-gen}
\noindent Extract all significant objects (nouns) from the following text prompt.

\noindent Text Prompt: \texttt{"\{prompt\}}

\begin{itemize}
    \item Identify all significant objects (nouns) mentioned in the prompt, including:
    \begin{itemize}
        \item Living beings (people, animals, creatures)
        \item Substantial physical objects (vehicles, furniture, buildings, tools, toys)
        \item Natural features (trees, rocks, mountains, water bodies)
        \item Important branded items or recognizable objects
    \end{itemize}
    \item Do \textbf{not} include:
    \begin{itemize}
        \item Positional/directional terms (\textit{top, bottom, left, right, front, back, side, center})
        \item Clothing items (shirts, pants, shoes, hats) unless they are the main focus
        \item Body parts (hands, face, legs) unless they are the main focus
        \item Abstract concepts (\textit{time, love, happiness})
        \item Very small or insignificant items (\textit{buttons, zippers, laces})
        \item Prepositions or spatial relationships (\textit{on, in, under, above, below})
    \end{itemize}
    \item Return the result as a JSON array containing object names in \textit{lowercase, singular form}.
\end{itemize}

\noindent For example:
\begin{itemize}
    \item \texttt{"a red car and blue bike"} $\rightarrow$ [\texttt{"car", "bike"}]
    \item \texttt{"large wooden table with metal legs"} $\rightarrow$ [\texttt{"table"}]
    \item \texttt{"small white dog running in the park"} $\rightarrow$ [\texttt{"dog", "park"}]
    \item \texttt{"three books on the shelf"} $\rightarrow$ [\texttt{"book", "shelf"}]
\end{itemize}

\end{PromptBox}

\begin{PromptBox}[]{Prompt 2: Used for spatial relationship}
\label{prompt:spat-rel}
\noindent Extract the spatial relationship and objects from the following text prompt.

\noindent Text Prompt: \texttt{"\{prompt\}}

\noindent Available spatial relationships: \texttt{\{available\_relationships\}}

\begin{itemize}
    \item Identify:
    \begin{itemize}
        \item The first object (\texttt{obj1}) - the reference object
        \item The spatial relationship - must be exactly one from the available relationships
        \item The second object (\texttt{obj2}) - the object being positioned relative to obj1
    \end{itemize}
    \item Return the result as a JSON object with exactly these keys:
    \begin{itemize}
        \item \texttt{"obj1"}: the first/reference object (lowercase, singular form)
        \item \texttt{"relationship"}: the spatial relationship (exactly as written in available relationships)
        \item \texttt{"obj2"}: the second object (lowercase, singular form)
    \end{itemize}
    \item If no valid spatial relationship is found, return an empty object: \texttt{\{\}}
    \item Focus on the main spatial relationship in the prompt
\end{itemize}

\noindent For example:
\begin{itemize}
    \item \texttt{"a red car on the left of a blue bike"} $\rightarrow$ \{\texttt{"obj1": "car", "relationship": "on the left of", "obj2": "bike"}\}
    \item \texttt{"dog next to the table"} $\rightarrow$ \{\texttt{"obj1": "dog", "relationship": "next to", "obj2": "table"}\}
    \item \texttt{"cat on top of the chair"} $\rightarrow$ \{\texttt{"obj1": "cat", "relationship": "on top of", "obj2": "chair"}\}
\end{itemize}

\end{PromptBox}

\begin{PromptBox}[]{Prompt 3: Used for object-attribute generation}
\label{prompt:attr-gen}
\noindent Extract objects and their attributes from the following text prompt.

\noindent Text Prompt: \texttt{"\{prompt\}}

\begin{itemize}
    \item Identify all objects (nouns) and their associated attributes (adjectives or descriptive words).
    \item Return the result as a JSON object where:
    \begin{itemize}
        \item Keys are object names (\textit{lowercase, singular form})
        \item Values are lists of attributes associated with each object
    \end{itemize}
    \item Only include objects that have explicit attributes mentioned.
    \item If an object has no attributes, don't include it.
\end{itemize}

\noindent For example:
\begin{itemize}
    \item \texttt{"a red car and blue bike"} $\rightarrow$ \{\texttt{"car": ["red"], "bike": ["blue"]}\}
    \item \texttt{"large wooden table"} $\rightarrow$ \{\texttt{"table": ["large", "wooden"]}\}
    \item \texttt{"small white dog running"} $\rightarrow$ \{\texttt{"dog": ["small", "white"]}\}
\end{itemize}

\end{PromptBox}

\begin{PromptBox}[]{Prompt 4: Used for generating numeracy}
\label{prompt:num-gen}
\noindent Extract all objects and their exact quantities from the following text prompt.

\noindent Text Prompt: \texttt{"\{prompt\}}

\noindent Available number words: \texttt{\{available\_numbers\}}

\begin{itemize}
    \item Identify all objects (nouns) and their associated quantities (numbers).
    \item Include both written numbers (\textit{e.g., "seven"}) and numeric digits (\textit{e.g., "7"}).
    \item Return a JSON object with exactly these keys:
    \begin{itemize}
        \item \texttt{"objects"}: a dictionary mapping object names to their quantities as integers
        \item \texttt{"numbers\_found"}: a list of the number words/digits found in the prompt
    \end{itemize}
\end{itemize}

\noindent For example:
\begin{itemize}
    \item \texttt{"seven horses and 4 pigs"} $\rightarrow$ \{\texttt{"objects": \{"horse": 7, "pig": 4\}, "numbers\_found": ["seven", "4"]}\}
    \item \texttt{"three red cars and two blue bikes"} $\rightarrow$ \{\texttt{"objects": \{"car": 3, "bike": 2\}, "numbers\_found": ["three", "two"]}\}
    \item \texttt{"a single dog and five cats"} $\rightarrow$ \{\texttt{"objects": \{"dog": 1, "cat": 5\}, "numbers\_found": ["a", "five"]}\}
\end{itemize}

\noindent Conversion rules:
\begin{itemize}
    \item "one"/"a"/"an"/"single" = 1
    \item "two"/"couple" = 2
    \item "three" = 3
    \item "four" = 4
    \item "five" = 5
    \item "six" = 6
    \item "seven" = 7
    \item "eight" = 8
    \item "nine" = 9
    \item "ten" = 10
    \item "dozen" = 12
    \item "hundred" = 100
\end{itemize}

\noindent Guidelines:
\begin{itemize}
    \item Convert object names to \textit{lowercase, singular form}.
    \item Only include objects that have explicit or implied quantities.
    \item If no specific number is mentioned but the object appears in a counting context, assume quantity 1.
    \item Focus on the main objects being counted in the prompt.
\end{itemize}

\end{PromptBox}

\begin{PromptBox}[]{Prompt 5: Used for generating synonyms}
\label{prompt:syn-gen}
\noindent Analyze the following list of objects and identify any synonyms or similar objects that should be grouped together.

\noindent Objects: \texttt{\{object\_names\}}

\begin{itemize}
    \item For each group of synonymous objects, choose the most common/standard term to represent the group.
    \item Return the result as a JSON object where:
    \begin{itemize}
        \item Keys are the standard/representative terms (\textit{lowercase, singular form}).
        \item Values are arrays of all synonymous terms that should be grouped under that key.
    \end{itemize}
\end{itemize}

\noindent For example:
\begin{itemize}
    \item \texttt{["boy", "child", "person", "girl"]} $\rightarrow$ \{\texttt{"person": ["boy", "child", "person", "girl"]}\}
    \item \texttt{["house", "home", "building"]} $\rightarrow$ \{\texttt{"house": ["house", "home", "building"]}\}
    \item \texttt{["car", "vehicle", "automobile"]} $\rightarrow$ \{\texttt{"car": ["car", "vehicle", "automobile"]}\}
    \item \texttt{["dog", "cat", "bird"]} $\rightarrow$ \{\texttt{"dog": ["dog"], "cat": ["cat"], "bird": ["bird"]}\}
    \item \texttt{["bee", "bird", "animal"]} $\rightarrow$ \{\texttt{"bee": ["bee"], "bird": ["bird", "animal"]}\}
    \item \texttt{["desk", "table", "furniture"]} $\rightarrow$ \{\texttt{"desk": ["desk", "table"], "furniture": ["furniture"]}\}
\end{itemize}

\noindent Only group objects that are truly synonymous or refer to the same type of thing. Keep unrelated objects separate.

\end{PromptBox}

\begin{PromptBox}[]{Prompt 6: Used for mapping synonyms}
\label{prompt:syn-map}
\noindent Map each detected label to at most one expected object based on semantic similarity, using the provided synonym groups as guidance.

\noindent Detected Labels: \texttt{\{detected\_labels\}}

\noindent Expected Objects (with synonym groups): \texttt{\{expected\_synonym\_groups\}}

\begin{itemize}
    \item For each detected label, decide whether it semantically refers to one of the expected objects (directly or via its synonym group).
    \item Enforce \textbf{one-to-one matching}: each detected label maps to \textit{at most one} expected object, and each expected object is the target of \textit{at most one} detected label. This prevents conflating similar but distinct objects (\textit{e.g.}, \texttt{stool} vs.\ \texttt{bench}, \texttt{cup} vs.\ \texttt{mug}).
    \item If a detected label has no good match among the expected objects, map it to \texttt{"none"}.
    \item Return the result as a JSON object where keys are detected labels (\textit{lowercase, singular form}) and values are the matched expected object or \texttt{"none"}.
\end{itemize}

\noindent For example:
\begin{itemize}
    \item Detected: \texttt{["bird", "tree"]}, Expected: \{\texttt{"albatross": ["albatross", "seabird", "large bird"], "oak": ["oak", "tree"]}\} $\rightarrow$ \{\texttt{"bird": "albatross", "tree": "oak"}\}
    \item Detected: \texttt{["stool", "bench"]}, Expected: \{\texttt{"bench": ["bench", "seat"]}\} $\rightarrow$ \{\texttt{"stool": "none", "bench": "bench"}\}
    \item Detected: \texttt{["puppy", "kitten", "cat"]}, Expected: \{\texttt{"dog": ["dog", "puppy", "canine"], "cat": ["cat", "kitten", "feline"]}\} $\rightarrow$ \{\texttt{"puppy": "dog", "kitten": "cat", "cat": "none"}\}
\end{itemize}

\noindent Only assign a match when the detected label clearly corresponds to an expected object or one of its synonyms; otherwise return \texttt{"none"} to avoid spurious matches.

\end{PromptBox}

\begin{PromptBox}[]{Prompt 7: VLM Prompt for criteria evaluation}
\label{prompt:crit-eval}
\begin{itemize}
    \item \textbf{Color}: \texttt{"What color is the \{object\}?"}
    \item \textbf{Shape}: \texttt{"What shape is the \{object\}?"}
    \item \textbf{Texture}: \texttt{"What texture is the \{object\}?"}
\end{itemize}

\end{PromptBox}

\begin{PromptBox}[]{Prompt 8: Used for evaluating attribute alignment}
\label{prompt:attr-align}
\noindent Use LLM to evaluate how well the response matches the expected attribute.

\noindent Args:
\begin{itemize}
    \item \texttt{response}: VLM response describing the object
    \item \texttt{expected\_attribute}: The attribute we're looking for
\end{itemize}

\noindent Returns: Score between 0.0 and 1.0

\noindent Prompt:
\begin{quote}
You are evaluating whether a description matches a specific attribute.

Expected attribute: "\{expected\_attribute\}" \\
Description: "\{response\}"

Rate how well the description matches the expected attribute on a scale from 0.0 to 1.0:
\begin{itemize}
    \item 1.0: Perfect match (the description clearly contains or describes the expected attribute)
    \item 0.8--0.9: Very good match (the description strongly suggests the expected attribute)
    \item 0.5--0.7: Partial match (the description somewhat relates to the expected attribute)
    \item 0.2--0.4: Weak match (there might be some connection but it's unclear)
    \item 0.0--0.1: No match (the description doesn't relate to the expected attribute at all)
\end{itemize}

Consider semantic similarity, synonyms, and related concepts. For example:
\begin{itemize}
    \item "red" matches "crimson" or "scarlet" (0.8--0.9)
    \item "large" matches "big" or "huge" (0.8--0.9)
    \item "wooden" matches "made of wood" (1.0)
    \item "smooth" might partially match "polished" (0.6--0.7)
\end{itemize}

Respond with only a number between 0.0 and 1.0 (e.g., "0.8").
\end{quote}

\end{PromptBox}

\section{Cost Analysis}

Using "SANEval" through basic hardware incurs task-specific API cost per image. Reported local CPU timings and costs appear in the detailed breakdown in Table~\ref{cost:breakdown}; these are not fixed end-to-end API wall-time estimates.

\begin{table}[h]
    \centering
    \caption{Cost and Latency Breakdown.}
    \label{cost:breakdown}
    \begin{tabular}{lcc}
        \hline
        \textbf{Pipeline Component} & \textbf{Latency (Local CPU)} & \textbf{Cost (USD)} \\
        \hline
        Spatial Evaluation & 0.11 s & \$0.02 \\
        Numeracy Evaluation & 0.12 s & \$0.03 \\
        Attribute Evaluation & 0.26 s & \$0.01 \\
        \hline
    \end{tabular}
\end{table}

\section{Released Dataset Schema} \label{AS.schema}
The public release at \url{https://huggingface.co/datasets/saneval/saneval-release} exposes five fields per record (Table~\ref{tab:schema}). Together they uniquely identify a prompt--image pair and place it within the SANEval taxonomy by (model, split, compositional task), supporting automated rescoring with the released evaluation pipeline.

\begin{table}[ht!]
    \centering
    \caption{Schema of the public SANEval release. The dataset contains $\sim$$5{,}250$ prompts from SANEval-Simple and SANEval-Hard, with $31{,}500$ intended prompt--model combinations and $29{,}998$ released records after failed generations are excluded.}
    \label{tab:schema}
    \begin{adjustbox}{max width=\linewidth}
    \begin{tabular}{l l l p{7.5cm}}
        \toprule
        \textbf{Field} & \textbf{Type} & \textbf{Cardinality} & \textbf{Description} \\
        \midrule
        \texttt{prompt} & string & free-form (8--223 chars) & The natural-language prompt fed to the T2I model. \\
        \texttt{model}  & string (categorical) & 6 values & Identifier of the originating T2I model: Imagen 3.0, Imagen 4.0, Imagen 4.0 Ultra, Nano Banana, Seedream 3.0, GPT Image 1. \\
        \texttt{split}  & string (categorical) & 2 values & \texttt{simple} (algorithmically generated, human-validated) or \texttt{hard} (fully human-written). \\
        \texttt{type}   & string (categorical) & 5 values & Compositional task: \texttt{spatial}, \texttt{numeracy}, \texttt{color}, \texttt{shape}, or \texttt{texture}. The latter three constitute the attribute-binding subset. \\
        \texttt{image}  & image & variable & The generated PNG/JPEG image (typical width $\approx$ $1.0$--$1.5$\,k px). \\
        \bottomrule
    \end{tabular}
    \end{adjustbox}
\end{table}

The released dataset deliberately ships only raw artifacts (prompts, images, and structural labels), not derived ones (conformity feedback, scorer outputs, human ratings). This separation lets users (i) repeat automated scoring on the released images, (ii) substitute their own scorers or detectors against the same prompt--image pairs, and (iii) extend the dataset with new T2I models without entangling raw data with one specific evaluation choice. The evaluation pipeline that consumes this schema is released at \url{https://github.com/nielseni6/saneval}.

\section{Spearman Correlation with CompBench++} \label{AS.spearman}
We complement the human-correlation analysis (Section~\ref{S:Results:Human}) with a Spearman rank correlation between SANEval and CompBench++ scores at the prompt-image level. Each cell of Table~\ref{tab:correlation_spearman} reports $\rho$ and nominal $p$ for one (model, category) pair. The reported Bonferroni correction covers $35$ tests at family-wise $\alpha=0.05$ (per-test threshold $\alpha'\!\approx\!1.43\times 10^{-3}$). All cells with $p<10^{-3}$ remain significant after correction; several cells fail the corrected threshold. Estimated $\rho$ values are uniformly small in magnitude ($|\rho|\!\le\!0.25$), indicating a weak monotonic relationship. This is consistent with SANEval and CompBench++ measuring related but non-equivalent signals; it does not establish strong human agreement with either evaluator (Section~\ref{S:Results:Human}).

\begin{table}[ht!]
    \centering
    \caption{Spearman rank correlation between SANEval and CompBench++ scores at the prompt-image level (originally reported coefficients and nominal p-values).}
    \label{tab:correlation_spearman}
    \begin{adjustbox}{max width=\linewidth}
    \begin{tabular}{l|c|c|ccc}
        \toprule
        \multirow{2}{*}{Model} & \multirow{2}{*}{Spatial} & \multirow{2}{*}{Numeracy}
        & \multicolumn{3}{c}{Attribute-Binding} \\
        \cmidrule(lr){4-6}
        & & & Color & Shape & Texture \\
        \midrule
        Imagen 3.0       & $\rho{=}{-}0.24$, $p<10^{-20}$ & $\rho{=}{-}0.19$, $p<10^{-20}$ & $\rho{=}{-}0.12$, $p<10^{-13}$ & \textit{$\rho{=}0.04$, $p{=}0.143$} & $\rho{=}{-}0.09$, $p{=}6.3\times 10^{-3}$ \\
        Imagen 4.0       & $\rho{=}{-}0.22$, $p<10^{-20}$ & $\rho{=}{-}0.18$, $p<10^{-20}$ & $\rho{=}{-}0.13$, $p<10^{-13}$ & $\rho{=}{-}0.10$, $p{=}1.7\times 10^{-3}$ & $\rho{=}{-}0.18$, $p{=}1.9\times 10^{-8}$ \\
        Imagen 4.0 Ultra & $\rho{=}{-}0.20$, $p<10^{-20}$ & $\rho{=}{-}0.14$, $p<10^{-20}$ & $\rho{=}{-}0.13$, $p<10^{-13}$ & $\rho{=}0.06$, $p{=}6.6\times 10^{-2}$ & $\rho{=}{-}0.09$, $p{=}5.8\times 10^{-3}$ \\
        Nano Banana      & $\rho{=}{-}0.20$, $p<10^{-20}$ & $\rho{=}{-}0.18$, $p<10^{-20}$ & $\rho{=}{-}0.15$, $p<10^{-15}$ & $\rho{=}{-}0.09$, $p{=}7.1\times 10^{-3}$ & $\rho{=}{-}0.18$, $p<10^{-7}$ \\
        Seedream 3.0     & $\rho{=}{-}0.21$, $p<10^{-20}$ & $\rho{=}{-}0.18$, $p<10^{-20}$ & $\rho{=}{-}0.16$, $p<10^{-15}$ & $\rho{=}{-}0.07$, $p{=}2.7\times 10^{-2}$ & $\rho{=}0.01$, $p{=}7.8\times 10^{-1}$ \\
        GPT Image 1      & $\rho{=}{-}0.20$, $p<10^{-20}$ & $\rho{=}{-}0.20$, $p<10^{-20}$ & $\rho{=}{-}0.11$, $p<10^{-10}$ & $\rho{=}{-}0.12$, $p{=}3.4\times 10^{-4}$ & $\rho{=}{-}0.10$, $p{=}1.4\times 10^{-3}$ \\
        \midrule
        \textbf{Reported summary} & $\boldsymbol{\rho{=}{-}0.21}$, $\boldsymbol{p<10^{-20}}$ & $\boldsymbol{\rho{=}{-}0.15}$, $\boldsymbol{p<10^{-20}}$ & $\boldsymbol{\rho{=}{-}0.11}$, $\boldsymbol{p<10^{-11}}$ & $\boldsymbol{\rho{=}{-}0.07}$, $\boldsymbol{p{=}3.1\times 10^{-2}}$ & $\boldsymbol{\rho{=}{-}0.05}$, $\boldsymbol{p{=}1.3\times 10^{-1}}$ \\
        \bottomrule
    \end{tabular}
    \end{adjustbox}
\end{table}

\section{Attribute Binding by Object Cardinality} \label{AS.cardinality}
Table~\ref{tab:attrbind} reports per-model attribute-binding scores stratified by the number of objects in the prompt ($1$, $2$, $3$--$5$, $6$--$10$, $>$$10$). The summary patterns reported in Section~\ref{S:Results} (shape brittleness, generally lower high-cardinality scores, fan-out at high cardinality) are read directly off this table.

\begin{table}[ht!]
    \centering
    \caption{Per-model SANEval attribute-binding scores (Color, Shape, Texture) as a function of object cardinality. Higher is better. Best score per column is in bold.}
    \label{tab:attrbind}
    \begin{adjustbox}{max width=\linewidth}
    \begin{tabular}{llcccccc}
        \toprule
        \textbf{Attribute} & \textbf{Model} & \textbf{Averaged} & \textbf{1 Obj} & \textbf{2 Obj} & \textbf{3--5 Obj} & \textbf{6--10 Obj} & \textbf{$>$10 Obj} \\
        \midrule
        \multirow{6}{*}{\textbf{Color}}
            & Imagen 3.0       & 0.5780 & 0.5913 & \textbf{0.6293} & 0.6013 & 0.5001 & 0.2548 \\
            & Imagen 4.0       & 0.5453 & 0.6013 & 0.5704 & 0.5178 & 0.4972 & 0.3364 \\
            & Imagen 4.0 Ultra & 0.6090 & 0.6597 & 0.6025 & 0.5965 & 0.5788 & 0.5037 \\
            & Nano Banana      & 0.5927 & \textbf{0.6667} & 0.6089 & 0.5911 & 0.4716 & 0.3890 \\
            & Seedream 3.0     & \textbf{0.6334} & 0.6343 & 0.6163 & \textbf{0.6647} & \textbf{0.6310} & \textbf{0.5244} \\
            & GPT Image 1      & 0.5850 & 0.6232 & 0.5885 & 0.5815 & 0.5427 & 0.4492 \\
        \midrule
        \multirow{6}{*}{\textbf{Shape}}
            & Imagen 3.0       & 0.2499 & 0.3159 & 0.2571 & 0.2394 & 0.1643 & 0.0612 \\
            & Imagen 4.0       & 0.2704 & 0.3647 & 0.2954 & 0.2229 & 0.1538 & 0.0957 \\
            & Imagen 4.0 Ultra & \textbf{0.3521} & \textbf{0.4311} & \textbf{0.3330} & \textbf{0.3339} & \textbf{0.2877} & \textbf{0.2195} \\
            & Nano Banana      & 0.2827 & 0.3649 & 0.2935 & 0.2306 & 0.1805 & 0.1198 \\
            & Seedream 3.0     & 0.2805 & 0.3195 & 0.2702 & 0.2810 & 0.2342 & 0.1997 \\
            & GPT Image 1      & 0.2972 & 0.4106 & 0.3094 & 0.2519 & 0.1650 & 0.1048 \\
        \midrule
        \multirow{6}{*}{\textbf{Texture}}
            & Imagen 3.0       & 0.4272 & 0.4090 & 0.4550 & 0.4108 & 0.4025 & 0.2836 \\
            & Imagen 4.0       & 0.4339 & 0.4663 & 0.4595 & 0.4035 & 0.3243 & 0.2353 \\
            & Imagen 4.0 Ultra & \textbf{0.5041} & 0.5471 & \textbf{0.5204} & 0.4610 & \textbf{0.4509} & \textbf{0.3589} \\
            & Nano Banana      & 0.4800 & \textbf{0.5731} & 0.4964 & 0.4142 & 0.3472 & 0.2767 \\
            & Seedream 3.0     & 0.4969 & 0.5442 & 0.4941 & \textbf{0.4836} & 0.4199 & 0.3246 \\
            & GPT Image 1      & 0.4655 & 0.4378 & 0.5158 & 0.4437 & 0.4094 & 0.3261 \\
        \bottomrule
    \end{tabular}
    \end{adjustbox}
\end{table}

\section{Pipeline Reliability Audit Protocol} \label{AS.audit}
The error-attribution table reported in Section~\ref{S:Results:Reliability} is derived from a manual audit of $24{,}012$ returned detections across $10{,}040$ image-level samples produced by the SANEval pipeline on SANEval-Simple. We document the audit protocol here.

\paragraph{Annotator pool.}
The audit was carried out by four expert annotators. Each annotator either holds a PhD, is actively pursuing a PhD in computer vision or machine learning, or has $10{+}$ years of professional experience in the AI industry. This composition was deliberate: distinguishing a true positive from a false positive in our pipeline often requires recognizing whether a near-synonym (e.g., a detected ``seabird'' standing in for a prompted ``albatross''), a partial occlusion, or an ambiguous boundary should count as a correct detection. These judgments benefit from domain expertise.

\paragraph{Annotation interface.}
Annotators used a purpose-built UI that displayed, side-by-side, (i) the structured pipeline log for each detection (the prompt's expected object, the prompt's expected attribute when applicable, the detector's raw label, the synonym group used as the query expansion, the synonym mapper's resolved expected-object assignment, and, for attribute-binding evaluations, the VLM's free-form attribute reading and the LLM-assigned similarity score) and (ii) the generated image with the candidate bounding box overlaid. Presenting the full log and the image jointly meant annotators could verify both the \emph{perceptual} claim (``is this region actually that object?'') and the \emph{LLM/VLM verification} claim (``did the downstream LLM/VLM components correctly map the detection to the right expected object and read the right attribute?'') in a single decision, without context-switching across pipeline stages.

\paragraph{Labeling rubric and two-stage attribution.}
For each detection, annotators recorded one of three labels: \texttt{true positive} (the bounding box localizes an instance that satisfies the prompt's expected object, and the downstream LLM/VLM verification correctly mapped the box to the right expected entry, including the right attribute reading when applicable); \texttt{false positive} (any failure that propagates to a wrong final score, regardless of where in the pipeline it originated); or \texttt{ambiguous} (genuinely unclear cases, e.g., partial occlusion below a discernibility threshold or stylistic abstraction). The reported binary tally contains $23{,}192$ valid and $820$ incorrect outputs; no separate ambiguous-case count is reported. For every false positive, annotators attributed the error to one of two pipeline stages: \emph{perception} (the open-vocabulary detector mislocalized or hallucinated the box, and no downstream LLM/VLM decision could have salvaged it), or \emph{LLM/VLM verification} (the perception output was correct, but a downstream LLM/VLM component caused the final-score failure: the synonym mapper assigned the box to the wrong expected object, the VLM-as-judge misread the attribute, or the LLM similarity scorer mis-graded the match against the expected attribute). Cases where neither stage alone explained the failure were attributed to \emph{both}. This two-stage decomposition is end-to-end with respect to the scored output: every false positive in the final SANEval score appears in exactly one of these three buckets.

\paragraph{Aggregation.}
The audit instances were partitioned across annotators with overlap on a held-out subset to assess agreement; final labels for non-overlap instances are individual judgments by one expert annotator, and overlap cases were adjudicated by majority vote with the tie-breaking convention of the more conservative (false-positive) label. The resulting $3.42\%$ incorrect-box fraction (Table~\ref{tab:fpr_attribution}) therefore reflects expert judgment under a per-detection, log-and-image rubric, not crowd-sourced or model-assisted labeling.

\section{Human Evaluation} \label{AS.humaneval}
\begin{table}[ht!]
    \centering
    \caption{Reported human-evaluation summary scores for spatial, numeracy, color, shape, and texture. Correlation targets use the original $1$--$5$ ratings, rather than these summaries.}
    \label{tab:humaneval}
    \begin{tabular}{l|c|c|ccc|c}
        \toprule
        \multirow{2}{*}{Model} & \multirow{2}{*}{Spatial} & \multirow{2}{*}{Numeracy}
        & \multicolumn{3}{c|}{Attribute-Binding} & \multirow{2}{*}{Total} \\
        \cmidrule(lr){4-6}
        & & & Color & Shape & Texture & \\
        \midrule
        Imagen 3.0       & 0.8160 & 0.7720 & 0.8400 & 0.5000 & 0.7240 & 0.7304\\
        Imagen 4.0       & 0.7800 & 0.6380 & 0.7990 & 0.5170 & 0.7410 & 0.6950\\
        Imagen 4.0 Ultra & \textbf{0.9390} & 0.8870 & \textbf{0.9680} & 0.8110 & \textbf{0.9190} & \textbf{0.9048}\\
        Nano Banana      & 0.9220 & 0.8680 & 0.9050 & 0.6810 & 0.8150 & 0.8382\\
        Seedream 3.0     & 0.9070 & 0.7430 & 0.9170 & 0.4310 & 0.6540 & 0.7304\\
        GPT Image 1      & 0.9250 & \textbf{0.8940} & \textbf{0.9680} & \textbf{0.8530} & 0.8430 & 0.8966\\
        \bottomrule
    \end{tabular}
\end{table}

Table~\ref{tab:humaneval} reports the per-category human-evaluation summary scores; the original $1$--$5$ ratings are used for the prompt-image-level Pearson correlations in Section~\ref{S:Results:Human}. Aggregated across all five categories, reported summary scores rank Imagen 4.0 Ultra highest, followed closely by GPT Image 1 and Nano Banana, with Imagen 4.0 lowest. This ordering differs from the SANEval ordering in Table~\ref{tab:combinedsaneval_diff}. Per-category differences are also informative: humans rate GPT Image 1 highest for numeracy and shape binding, where SANEval awards the lead to Imagen 4.0 Ultra. The prompt-image-level correlations assess agreement within each category ($r = 0.241$ for numeracy and $r = 0.212$ for shape), complementing these model-level summaries.

Table~\ref{tab:annotator_agreement} reports agreement on the original individual ratings, before rounding for the evaluator comparison. Ordinal and nominal $\alpha$ denote Krippendorff's agreement coefficients. ICC(A,1) and ICC(A,k) measure absolute agreement for a single rating and the mean of the four ratings, respectively. These statistics describe rating reliability and do not define a Pearson-correlation ceiling.

\begin{table}[ht!]
    \centering
    \caption{Inter-annotator agreement and mean human ratings by category. Four annotators independently rated $100$ prompt--image pairs per category ($500$ overall) on a $1$--$5$ Likert scale.}
    \label{tab:annotator_agreement}
    \begin{adjustbox}{max width=\linewidth}
    \begin{tabular}{lccccc}
        \toprule
        \textbf{Category} & \textbf{Mean rating} & \textbf{Ordinal $\alpha$} & \textbf{Nominal $\alpha$} & \textbf{ICC(A,1)} & \textbf{ICC(A,k)} \\
        \midrule
        Spatial & 3.933 & 0.830 & 0.550 & 0.897 & 0.972 \\
        Numeracy & 3.946 & 0.889 & 0.667 & 0.922 & 0.979 \\
        Color & 4.005 & 0.756 & 0.489 & 0.834 & 0.953 \\
        Shape & 3.767 & 0.880 & 0.603 & 0.881 & 0.967 \\
        Texture & 3.988 & 0.856 & 0.605 & 0.908 & 0.975 \\
        \midrule
        \textbf{Overall} & 3.928 & 0.848 & 0.585 & 0.890 & 0.970 \\
        \bottomrule
    \end{tabular}
    \end{adjustbox}
\end{table}

\begin{figure}[ht!]
\centering
\includegraphics[width=0.9\textwidth,keepaspectratio]{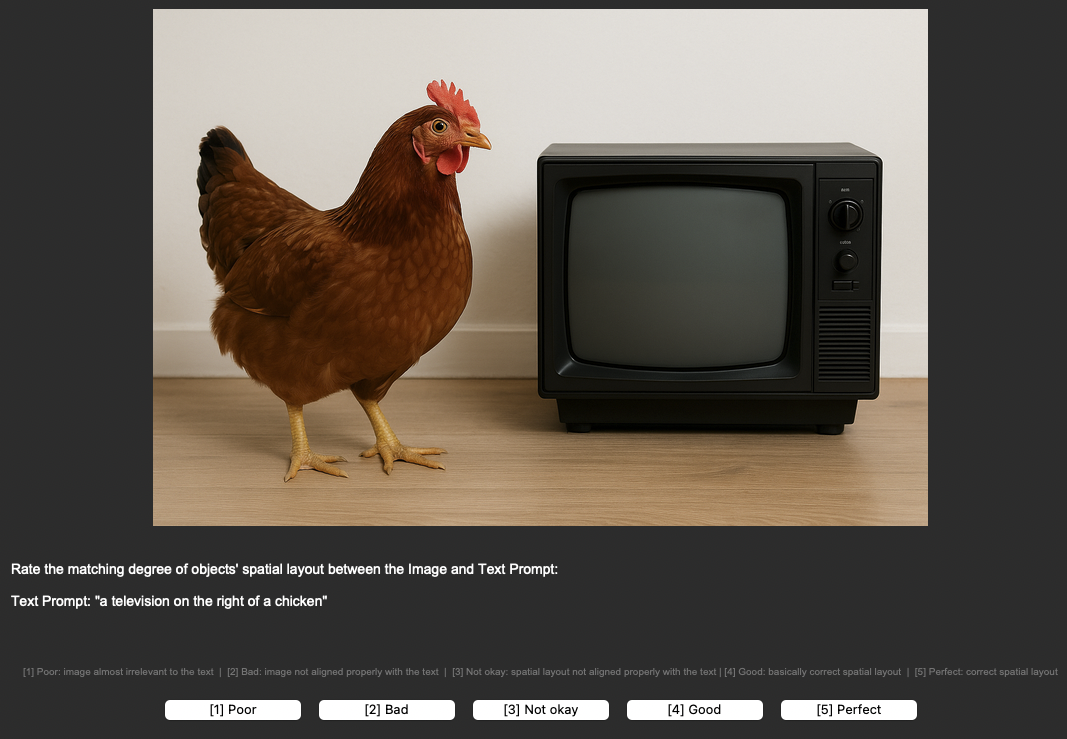}
\caption{Human evaluation interface used to collect adherence responses for spatial dataset.}
\label{fig:human_spatial}
\end{figure}

\begin{figure}[ht!]
\centering
\includegraphics[width=0.9\textwidth,keepaspectratio]{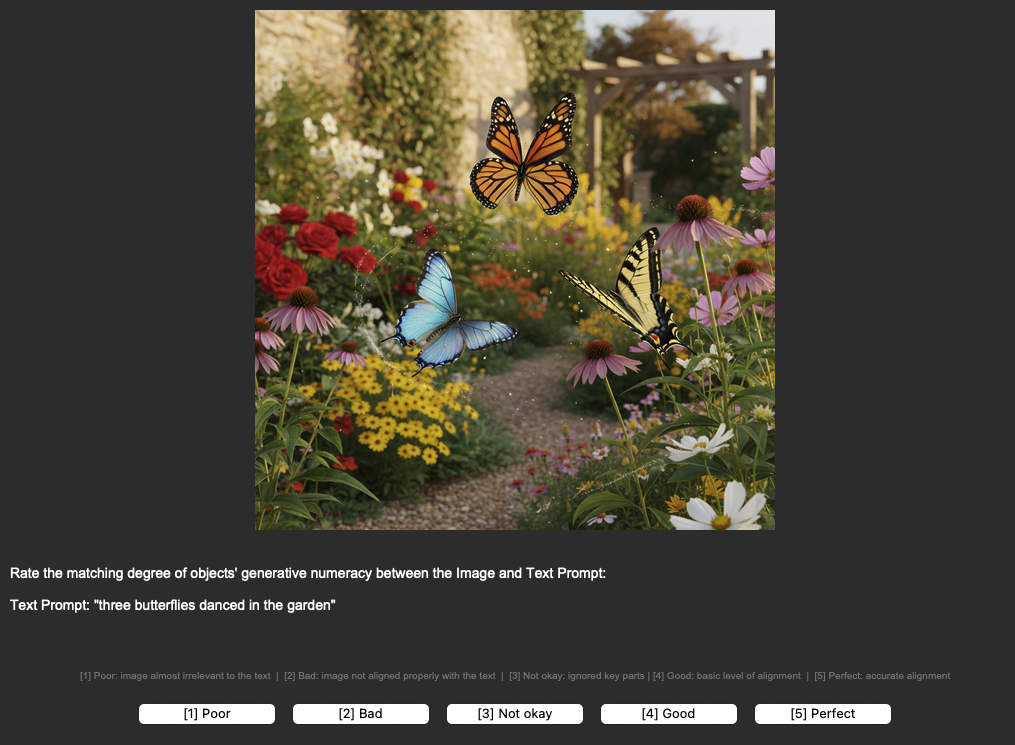}
\caption{Human evaluation interface used to collect adherence responses for numeracy dataset.}
\label{fig:human_numeracy}
\end{figure}

\begin{figure}[ht!]
\centering
\includegraphics[width=0.9\textwidth,keepaspectratio]{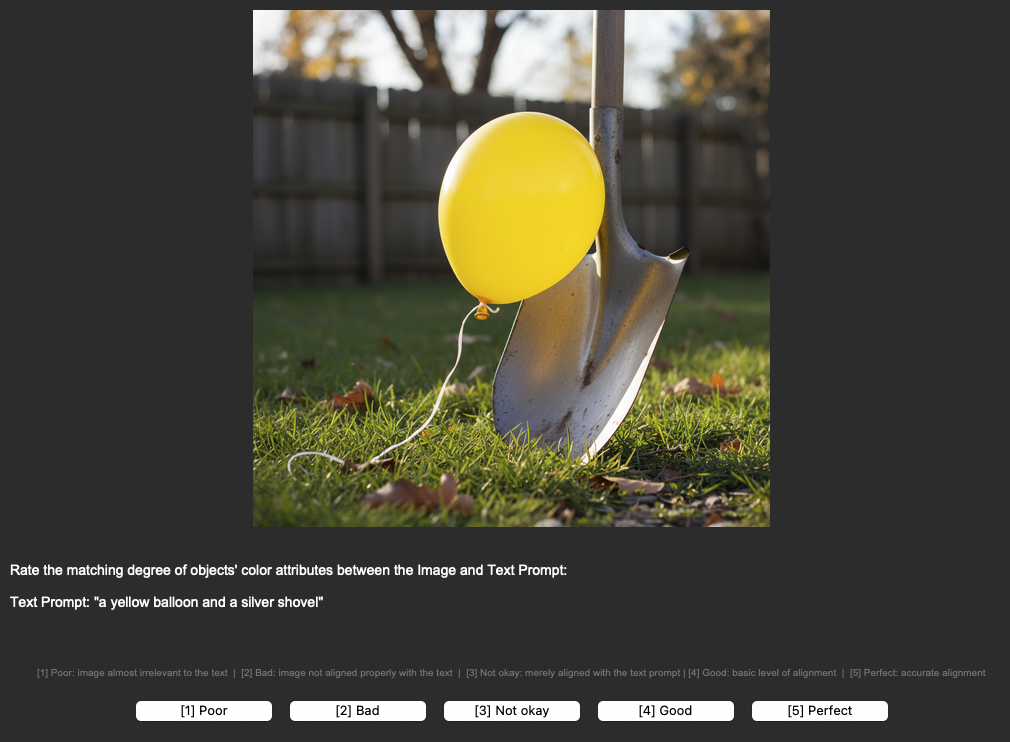}
\caption{Human evaluation interface used to collect adherence responses for color binding dataset.}
\label{fig:human_color}
\end{figure}

\begin{figure}[ht!]
\centering
\includegraphics[width=0.9\textwidth,keepaspectratio]{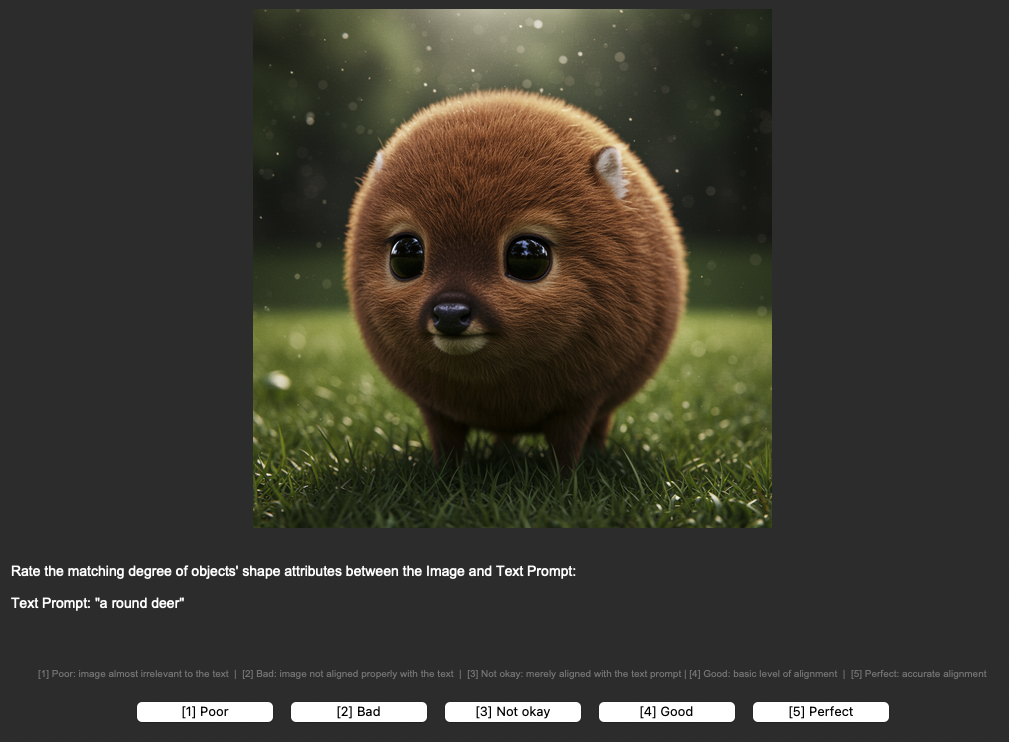}
\caption{Human evaluation interface used to collect adherence responses for shape binding dataset.}
\label{fig:human_shape}
\end{figure}

\begin{figure}[ht!]
\centering
\includegraphics[width=0.9\textwidth,keepaspectratio]{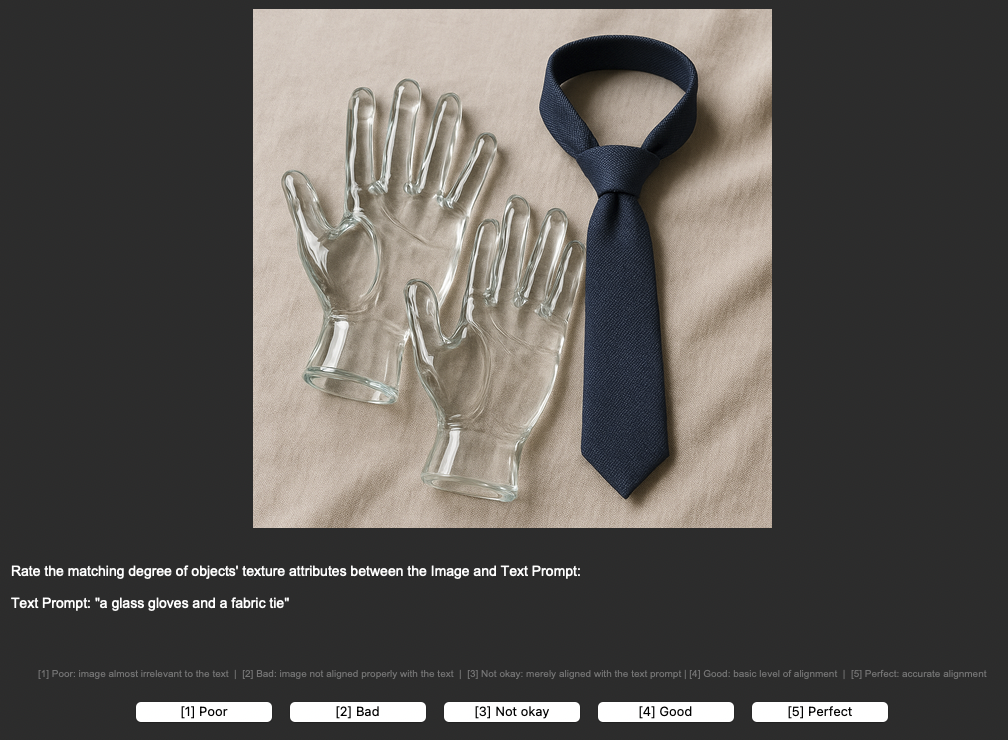}
\caption{Human evaluation interface used to collect adherence responses for texture binding dataset.}
\label{fig:human_texture}
\end{figure}

To validate the reliability of SANEval, we conducted a human evaluation study across all five categories of compositional adherence: spatial relationships, numeracy, and attribute binding (color, shape, and texture). The study was carried out by the same pool of four expert annotators used for the reliability audit (Appendix~\ref{AS.audit}): each annotator either holds a PhD, is actively pursuing a PhD in computer vision or machine learning, or has $10{+}$ years of professional experience in the AI industry. Using domain experts rather than crowdsourced workers was deliberate, since judging compositional adherence often hinges on subtle distinctions (e.g., whether a partially occluded object should count, or whether a near-shape qualifies as the requested geometry) that benefit from technical familiarity with how generative models tend to fail.

For each case, annotators were presented with a generated image and the corresponding text prompt, and asked to judge whether the specified conditions were satisfied on a $1$ (Poor) to $5$ (Perfect) Likert scale. The interface (Figures~\ref{fig:human_spatial}--\ref{fig:human_texture}) was uniform across tasks, with the prompt and candidate image displayed side-by-side and the Likert response options inline. Correlation targets are the rounded means of the four ratings, with an unrounded sensitivity analysis on the same in-vocabulary pairs.

\clearpage
\section{Qualitative Analysis}
Table~\ref{tab:qualitative_analysis} provides illustrative examples from the SANEval benchmark, highlighting how scores are assigned and how diagnostic feedback pinpoints specific failure modes across different evaluation categories.

\begin{table}[ht!]
    \centering
    \caption{Qualitative sample prompts, generated images, scores, and diagnostic feedback from SANEval. Attribute entries show detected-object binding scores; other entries show prompt scores.}
    \label{tab:qualitative_analysis}

    \renewcommand{\arraystretch}{1.5}
    \setlength{\tabcolsep}{6pt}

    \begin{tabular}{
        >{\centering\arraybackslash}m{1.7cm}
        >{\centering\arraybackslash}m{2.6cm}
        >{\centering\arraybackslash}m{2.4cm}
        >{\centering\arraybackslash}m{1.0cm}
        >{\centering\arraybackslash}m{4.1cm}
    }
        \toprule
        \textbf{Evaluation} & \textbf{Prompt} & \textbf{Image} & \textbf{Score} & \textbf{Conformity Feedback} \\
        \midrule
        Spatial & ``a mouse on the bottom of a painting''
        & \includegraphics[height=1.6cm]{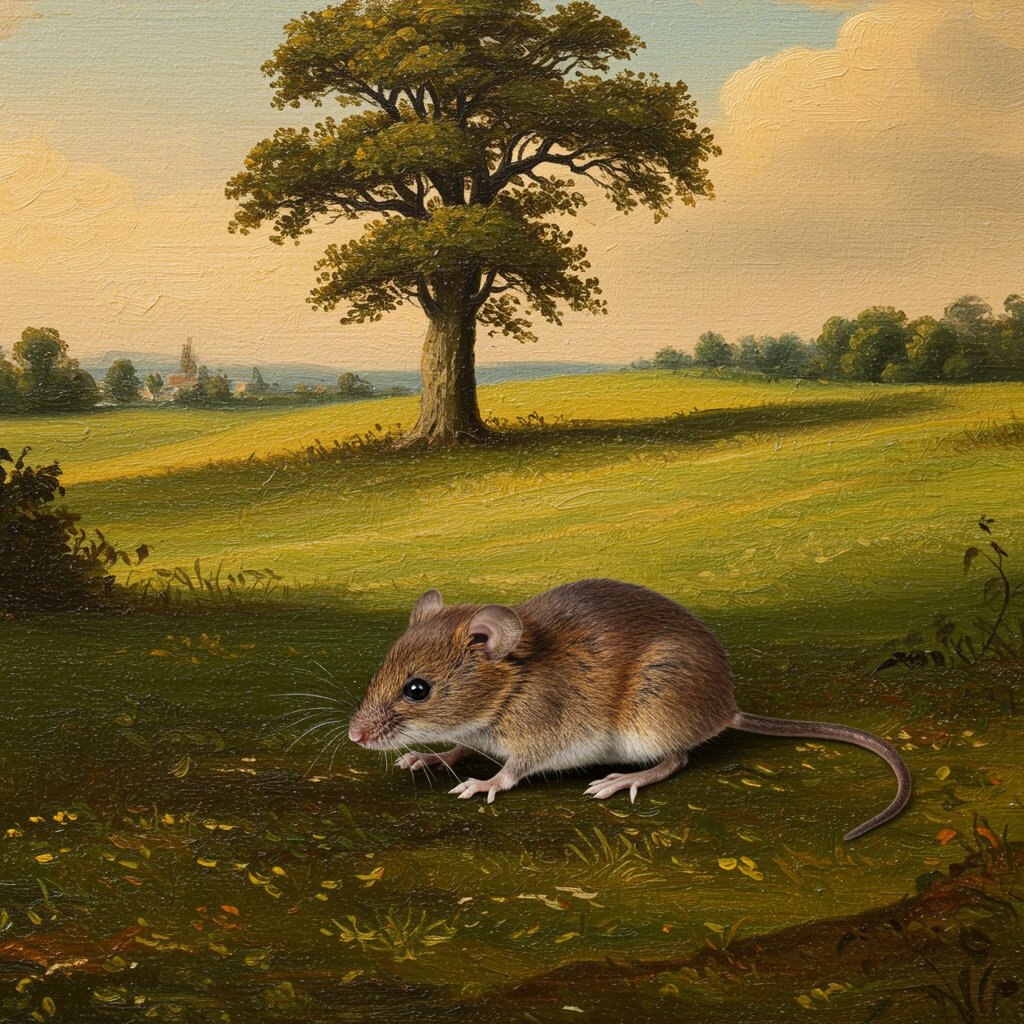}
        & 0.0
        & Missing object: painting \\

        Numeracy & ``four trucks''
        & \includegraphics[height=1.6cm]{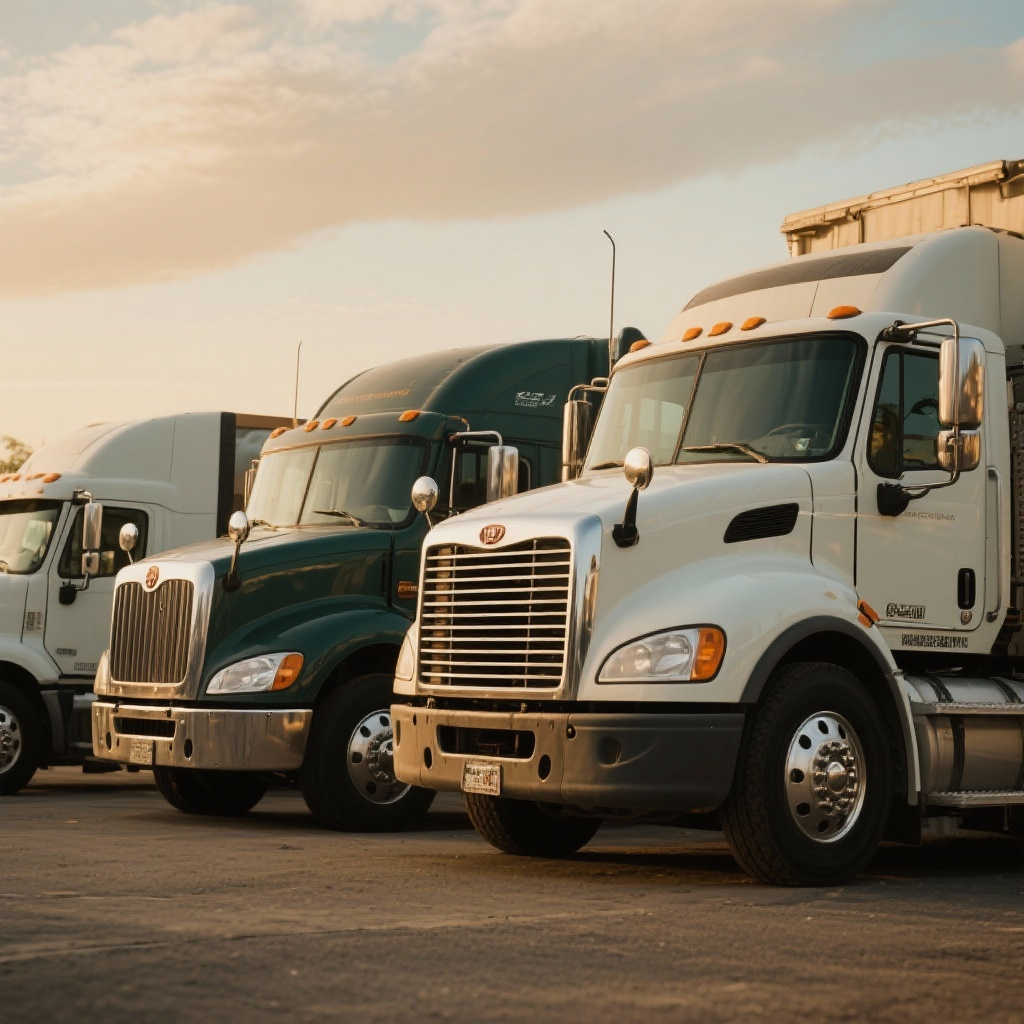}
        & 0.5
        & Two trucks missing \\

        Color Binding & ``a brown shirt and a pink apple''
        & \includegraphics[height=1.6cm]{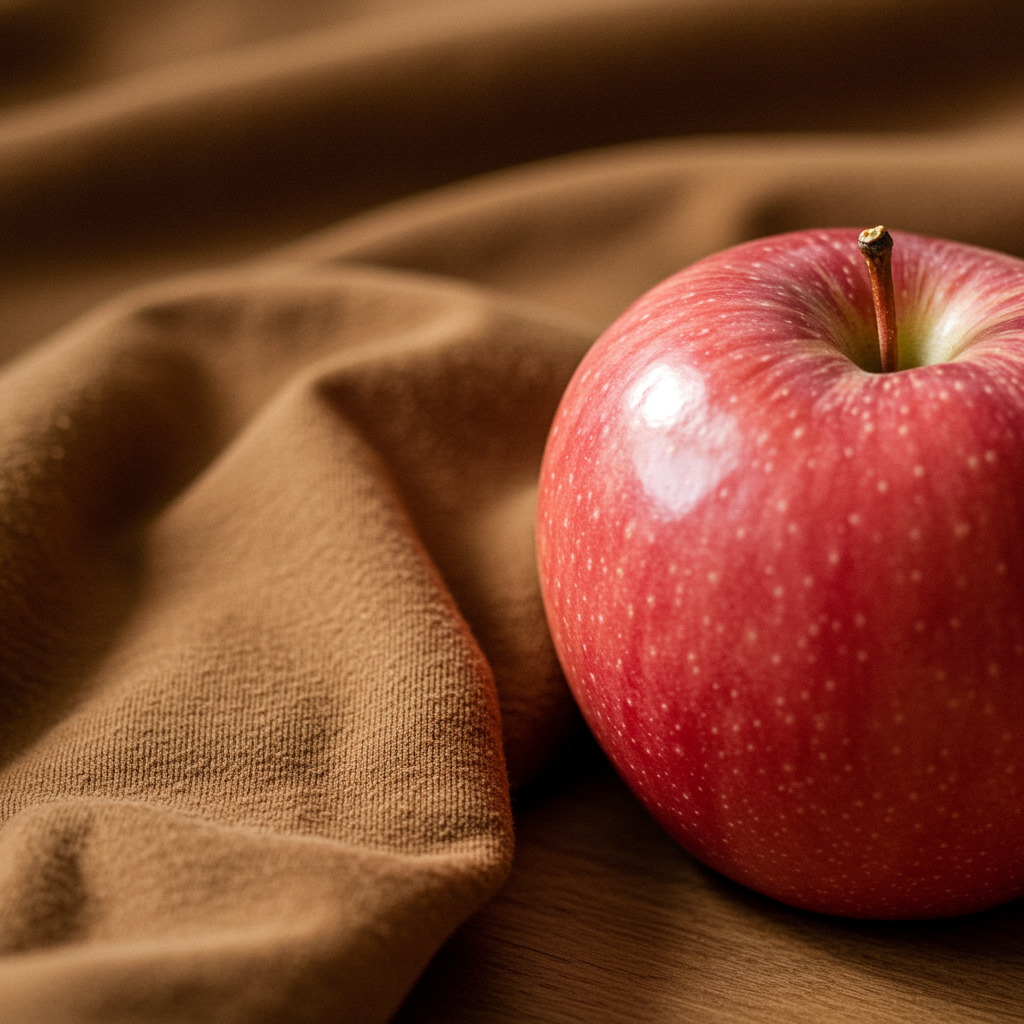}
        & 0.2
        & Missing object: shirt [brown]; Poor attribute binding for apple: expected [pink], score = 0.20 $<$ 0.75 \\

        Shape Binding & ``a pentagonal spoon''
        & \includegraphics[height=1.6cm]{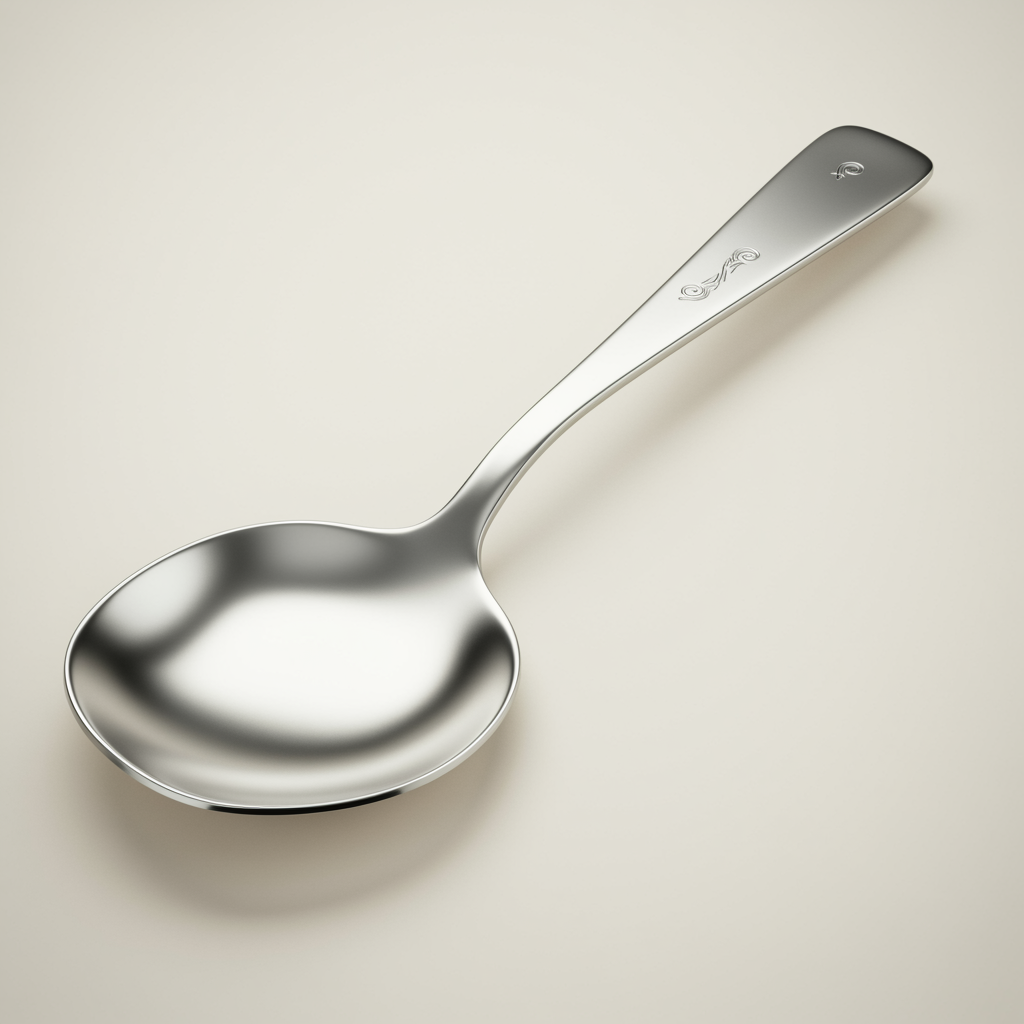}
        & 0.0
        & Poor attribute binding for spoon: expected [pentagonal], score = 0.00 $<$ 0.75 \\

        Texture Binding & ``a rubber grass''
        & \includegraphics[height=1.6cm]{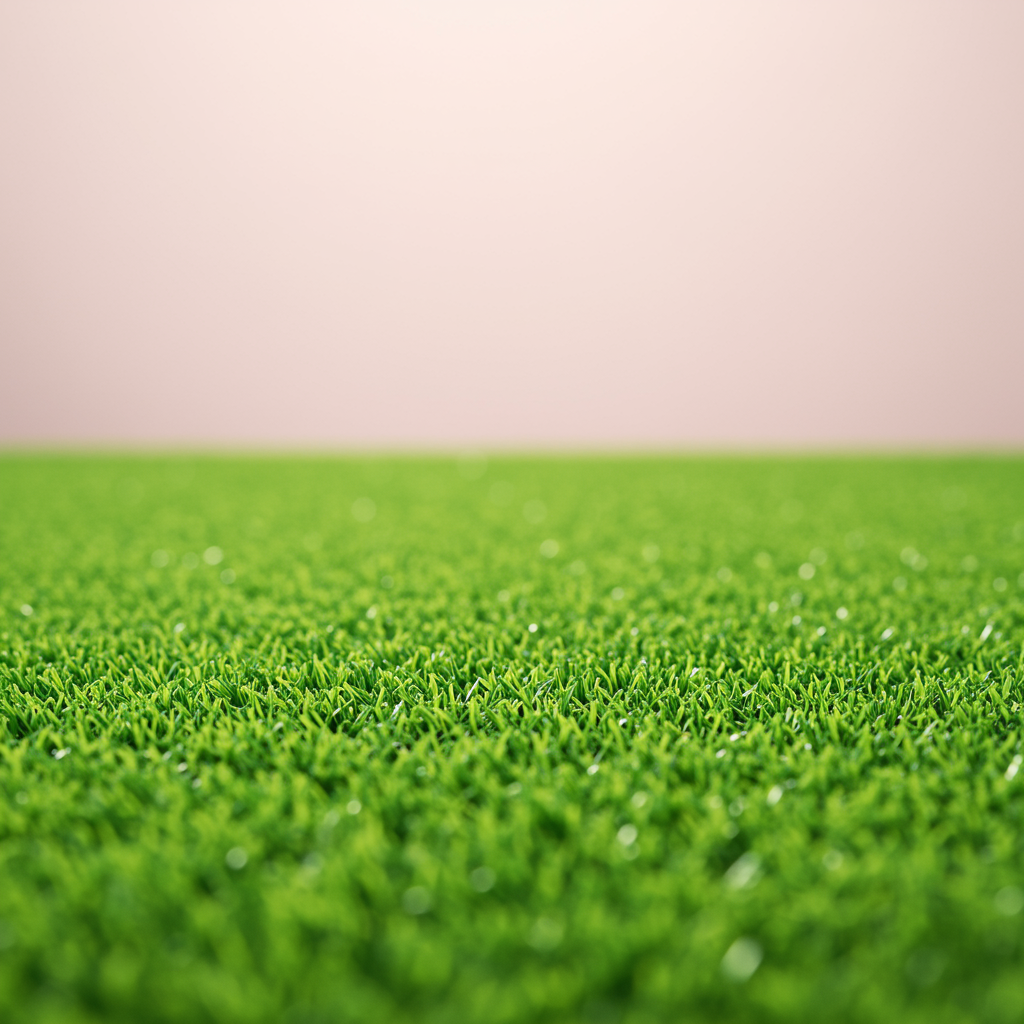}
        & 0.3
        & Poor attribute binding for grass: expected [rubber], score = 0.30 $<$ 0.75 \\
        \bottomrule
    \end{tabular}
\end{table}

These examples demonstrate SANEval's ability to go beyond scalar scores by producing structured, interpretable feedback (e.g., reporting missing objects or incorrect attributes), which makes model shortcomings explicit at a per-error level.

\subsection{Saneval Hard}
\begin{figure}[ht!]

        \includegraphics[width=0.9\textwidth,keepaspectratio]{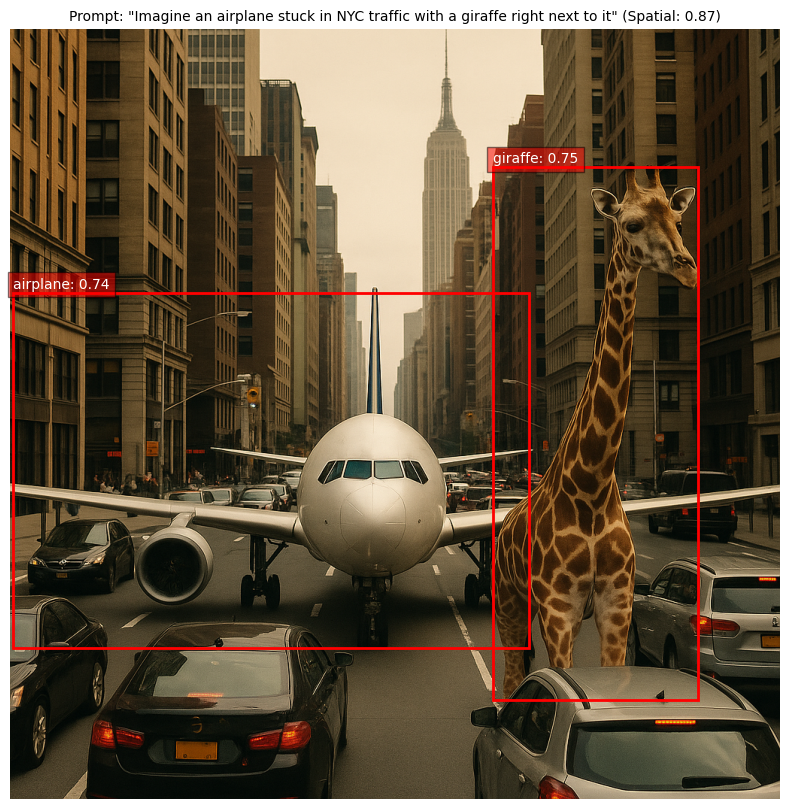}
        \caption{Examples from the SANEval-Hard prompt set: Spatial}
\end{figure}
\begin{figure}[ht!]

        \includegraphics[width=0.9\textwidth,keepaspectratio]{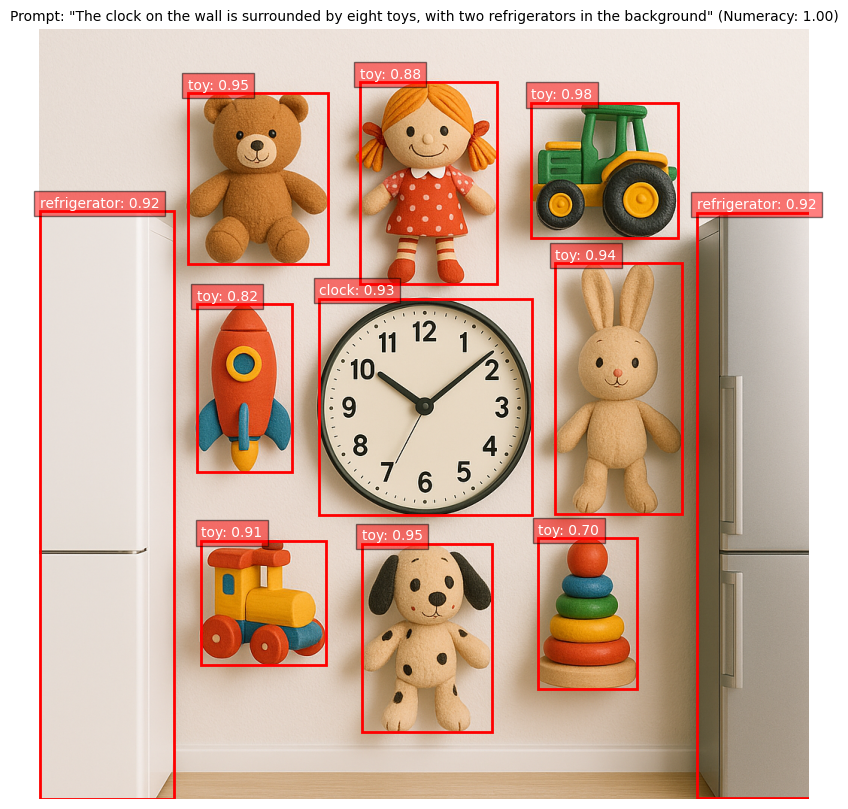}
   \caption{Examples from the SANEval-Hard prompt set: Numeracy}

\end{figure}
\begin{figure}[ht!]
\includegraphics[width=0.9\textwidth,keepaspectratio]{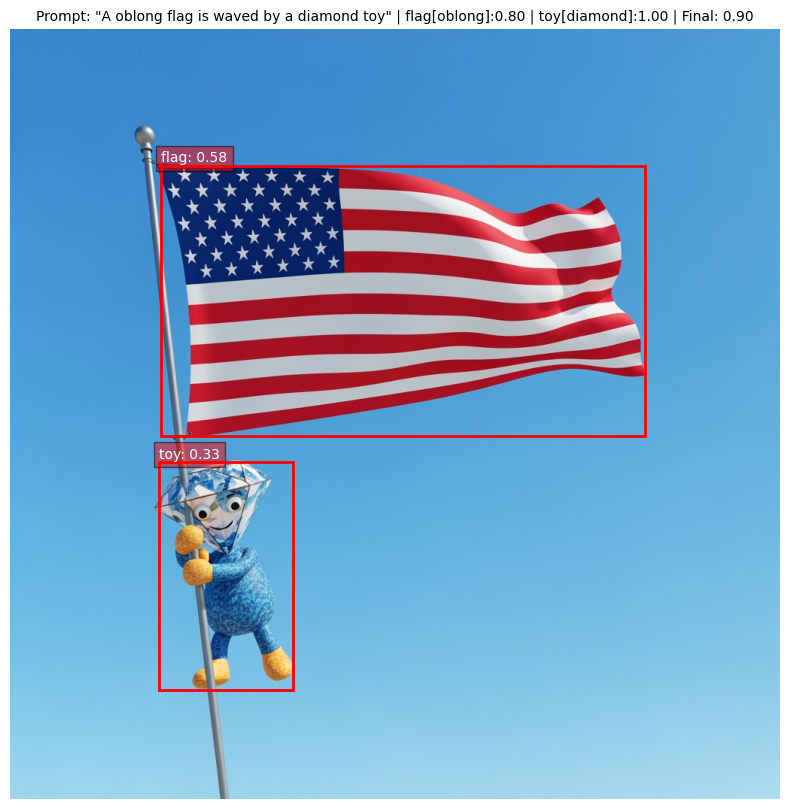}
    \caption{Examples from the SANEval-Hard prompt set: Attribute Binding}
    \label{fig:saneval_examples}
\end{figure}

\end{document}